\documentclass[10pt,twocolumn,letterpaper]{article}

\usepackage{iccv}
\usepackage{times}
\usepackage{epsfig}
\usepackage{graphicx}
\usepackage{dsfont}
\usepackage{amsmath}
\usepackage{mathrsfs}
\usepackage{amssymb}
\usepackage[numbers]{natbib}
\usepackage{multicol}
\usepackage{xparse} 
\usepackage[pagebackref=true,breaklinks=true,letterpaper=true,colorlinks,bookmarks=false]{hyperref}
\usepackage{cleveref}
\usepackage[font=small,position=b]{subcaption}
\usepackage{algorithmicx}
\usepackage{algorithm}
\usepackage[noend]{algpseudocode}
\usepackage{setspace}

\algrenewcommand\algorithmicrequire{\textbf{Input:}}
\algrenewcommand\algorithmicensure{\textbf{Output:}}
\algnewcommand\algorithmicinput{\textbf{Input:}}
\algnewcommand\INPUT{\item[\algorithmicinput]}

\usepackage{booktabs}
\usepackage[para]{threeparttable}
\usepackage{multirow}
\usepackage{pythonhighlight}

\NewDocumentCommand\bbm{}{ \begin{bmatrix} }
\NewDocumentCommand\ebm{}{ \end{bmatrix} }
\NewDocumentCommand\Vector{m}{ \boldsymbol{\mathbf{#1}} }
\NewDocumentCommand\Matrix{m}{ \boldsymbol{\mathbf{#1}} }
\NewDocumentCommand\Transpose{m}{ \left.{#1}\right.^T }

\NewDocumentCommand\Determinant{m}{ \mathrm{det}\left(#1\right) }
\NewDocumentCommand\Norm{m}{\left\Vert#1\right\Vert }

\NewDocumentCommand\ArgMin{m}{ \operatorname*{argmin}_{#1} }

\NewDocumentCommand\Real{}{ \mathbb{R} }

\NewDocumentCommand\LieGroupSO{m}{ \mathrm{SO}(#1) }

\NewDocumentCommand\NormalDistribution{mm}{ \mathcal{N}\left(#1,#2\right) }

\NewDocumentCommand\Rotation{}{ \Matrix{R} }

\NewDocumentCommand\Transform{}{ \Matrix{T} }

\NewDocumentCommand\Estimate{m}{\hat{#1}}

\NewDocumentCommand\Mean{m}{\overline{#1}}

\NewDocumentCommand\MatLog{m}{\mathrm{Log}\left({#1}\right)}
\NewDocumentCommand\MatExp{m}{\mathrm{Exp}\left({#1}\right)}

\NewDocumentCommand\RotDist{m}{d_\mathrm{#1}}
\NewDocumentCommand\quat{}{\Vector{q}}
\NewDocumentCommand\Covd{}{\Matrix{\Sigma}_a}
\NewDocumentCommand\Covh{}{\Matrix{\Sigma}_e}
\NewDocumentCommand\sigmad{}{\sigma_a}
\NewDocumentCommand\sigmah{}{\sigma_e}

\NewDocumentCommand\ImageCovariance{}{\Matrix{\Sigma_y}}

\graphicspath{{figs/}}

\iccvfinalcopy 

\def\iccvPaperID{2302} 

\ificcvfinal\fi
\begin{document}

\title{Probabilistic Regression of Rotations using Quaternion Averaging \\ and a Deep Multi-Headed Network}

\author{Valentin Peretroukhin\thanks{Corresponding author can be reached at {\tt\small v.peretroukhin@mail.utoronto.ca}.}, Brandon Wagstaff, Matthew Giamou, and Jonathan Kelly\\
University of Toronto\\
}

\maketitle

\begin{abstract}
Accurate estimates of rotation are crucial to vision-based motion estimation in augmented reality and robotics. In this work, we present a method to extract probabilistic estimates of rotation from deep regression models. First, we build on prior work and argue that a multi-headed network structure we name HydraNet provides better calibrated uncertainty estimates than methods that rely on stochastic forward passes. Second, we extend HydraNet to targets that belong to the rotation group, SO(3), by regressing unit quaternions and using the tools of rotation averaging and uncertainty injection onto the manifold to produce three-dimensional covariances. Finally, we present results and analysis on a synthetic dataset, learn consistent orientation estimates on the 7-Scenes dataset, and show how we can use our learned covariances to fuse deep estimates of relative orientation with classical stereo visual odometry to improve localization on the KITTI dataset.
\end{abstract}

\vspace{-1em}
\section{Introduction}
Accounting for position and orientation, or pose, is at the heart of computer vision. Many algorithms in image classification and feature tracking, for example, are explicitly concerned with output that is robust to camera orientation. Conversely, algorithms like visual odometry, structure from motion, and SLAM use visual sensors to estimate and track the pose of a camera as it moves through some environment. The algorithms in this latter category form the basis of visual localization pipelines in autonomous vehicles, aid in aerial vehicle navigation and mapping, and are often crucial to augmented reality applications.

Recent work  \cite{Clark2017, Melekhov2017-dl, Kendall2015-kh} has attempted to  transfer the success of deep neural networks in many areas of computer vision to the task of camera pose estimation. These approaches, however, can produce arbitrarily poor pose estimates if sensor data differs from what is observed during training (i.e., it is `out of training distribution') and their monolithic nature makes them difficult to debug. Further, despite much research effort, classical motion estimation algorithms, like stereo visual odometry, still achieve state-of-the-art performance in nominal conditions\footnote{Based on the KITTI odometry leaderboard \cite{Geiger2013-ky} at the time of writing.}. Nevertheless, the representational power of deep regression algorithms makes them an attractive option to complement classical motion estimation when these latter methods perform poorly (e.g., under diverse lighting conditions or low scene texture). By endowing deep regression models with a useful notion of uncertainty, we can account for out-of-training-distribution errors and fuse these models with classical methods using probabilistic factor graphs. In this work, we choose to focus on rotation regression, since many motion algorithms are sensitive to rotation errors \cite{Peretroukhin2018}, and good rotation initializations can be critical to robust optimization. Our novel contributions are
\begin{figure}
	\centering
	\includegraphics[width=0.48\textwidth]{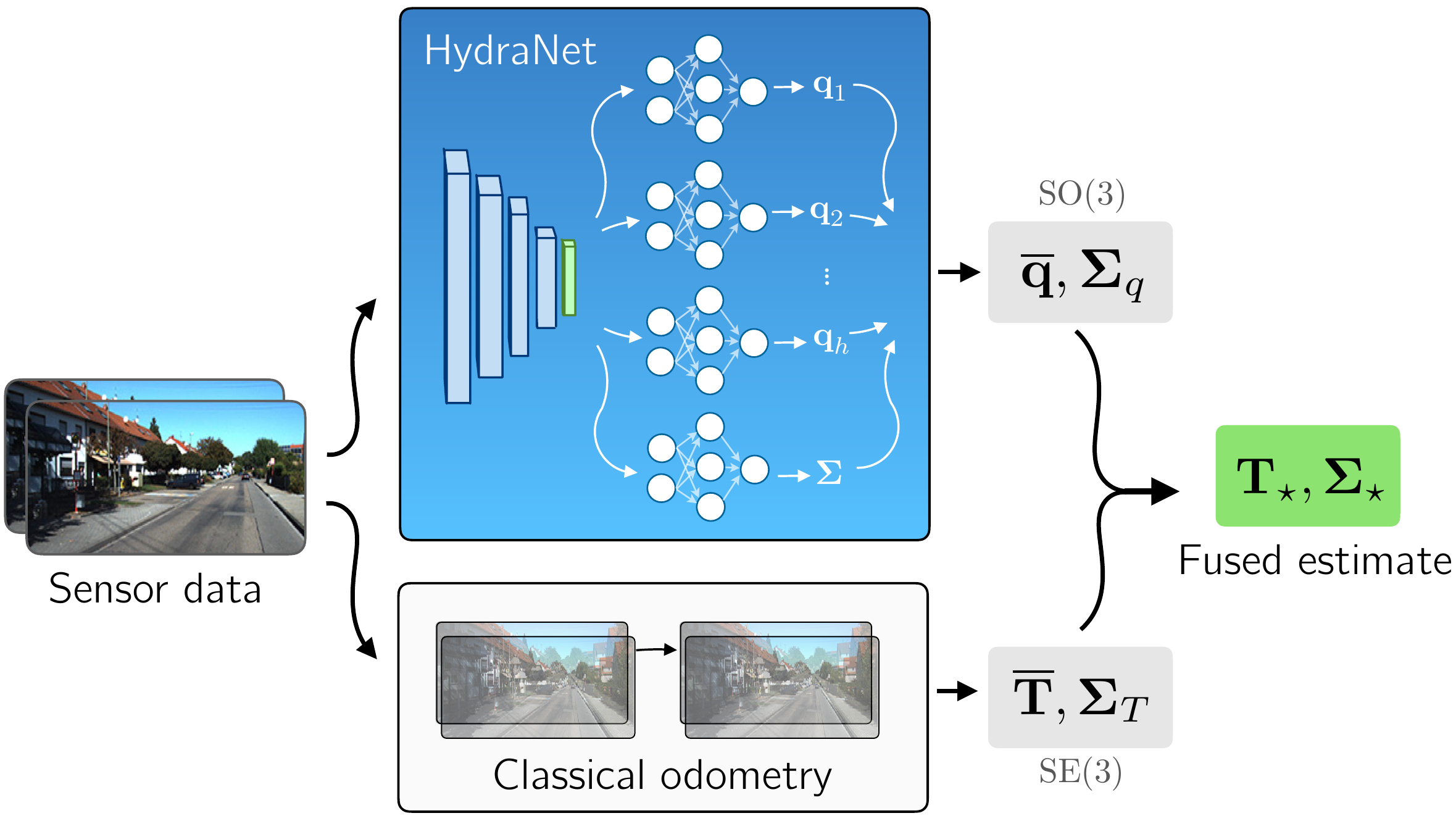}
	\caption{We improve classical pose estimation by fusing it with deep probabilistic models.}
	\label{fig:fusion_fig}
	\vspace{-1em}
\end{figure}
\vspace{-0.5em}
\begin{enumerate}
\item a deep network structure we call \textit{HydraNet} that builds on prior work \cite{Lakshminarayanan2017,Osband2016} to produce meaningful uncertainties over unconstrained targets,
\item a loss formulation and mathematical framework that extends HydraNet to means and covariances of the rotation group $\LieGroupSO{3}$,
\item and open source code for $\LieGroupSO{3}$ regression\footnote{Code will be released after the double-blind review process.}.
\end{enumerate}


\section{Related work}

%

Much recent work in the literature has been devoted to replacing classical localization algorithms with deep network equivalents. Some approaches \cite{Clark2017, Kendall2015-kh, Kendall2017-ud, Melekhov2017-dl} learn poses  directly, while others learn them indirectly as the spatial transforms that result in minimal loss defined over some other domain (e.g., pixel or depth space) \cite{Byravan2017-ik, Handa2016-hm}.

Despite this surge of research in neural-network-based replacements, some authors have nevertheless used deep networks to augment classical state estimation algorithms. Deep networks have been trained as pose correctors whose corrections can be fused with existing estimates through pose graph relaxation \cite{2018_Peretroukhin_Deep}, and as  depth prediction networks that can be incorporated into a classical monocular pipelines to provide an initial estimate for metric scale \cite{yang:2018}. Our work is perhaps closest in spirit to \cite{Haarnoja2016-ph} which fuses deep probabilistic observation functions with classical models using a Kalman Filter, but focuses on unconstrained targets and does not investigate uncertainty quantification.

In the robotics community, there has been significant effort to leverage the tools of matrix Lie groups to handle poses and associated uncertainty \cite{Sola2018-kg, Barfoot2014-ac}. In parallel, the computer vision community has developed a rich literature of rotation averaging \cite{Hartley2013-rc} which focuses on principled ways to combine elements of $\LieGroupSO{3}$ based on different metrics defined over the group.

Finally, ensembles of networks have been shown to be a scalable way to extract uncertainty for deep regression and classification  \cite{Lakshminarayanan2017}, while multi-headed networks have been proposed in the context of ensemble learning \cite{Lee2015-af} and for bootstrapped uncertainty in reinforcement learning \cite{Osband2016}. 

\section{Approach}

We develop our method for probabilistic $\LieGroupSO{3}$ regression in three steps. 
First, we motivate why learning elements of $\LieGroupSO{3}$ is particularly germane to field of egomotion estimation. Then, we present a multi-headed network that can regress unconstrained targets and produce consistent uncertainty estimates. Toward this end, we present a one-dimensional regression experiment, validating prior works \cite{Lakshminarayanan2017,Osband2016} that suggest a bootstrap-inspired approach provides better calibrated uncertainties than one based on stochastic sampling. Finally, we extend these results to targets that belong to $\LieGroupSO{3}$ by defining a rotation average using the quaternionic metric, and show how we can compute anisotropic uncertainty on four-dimensional unit quaternions.

\begin{figure}
	\centering
	\includegraphics[width=0.48\textwidth]{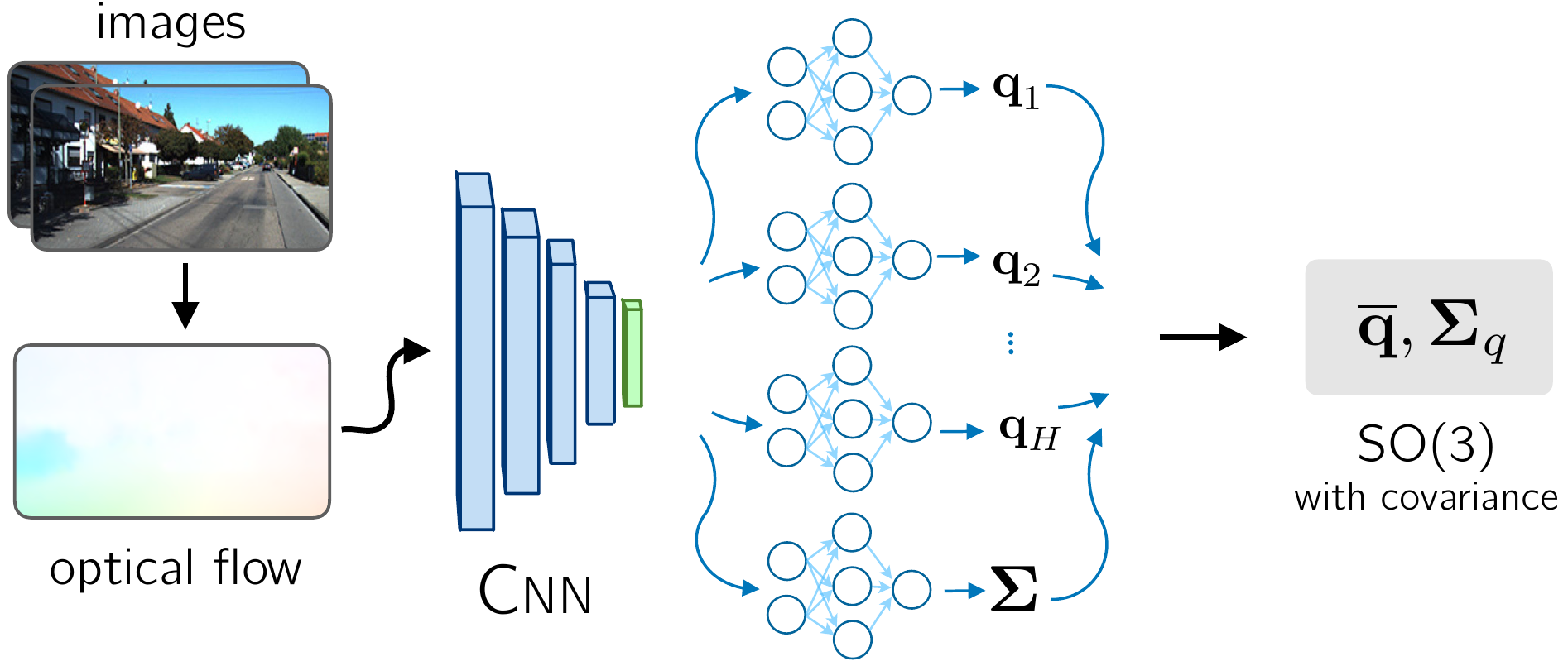}
	\caption{The HydraNet structure. Input data (in this case, pre-processed optical flow images) is passed through a main body and then through a number of heads. Outputs are combined to produce an average and an uncertainty.}
	\label{fig:kitti_flow_hydranet}
	\vspace{-1.5em}
\end{figure}

\subsection{Why Rotations?}
We focus our attention on learning rotations for a number of reasons. 
First, rotations can be learned without reference to scale, using monocular images without the need for metric depth estimation. These images can come from cheap, light-weight imaging sensors that can be found on many ground and aerial vehicles. Furthermore, many depth-equipped sensors like stereo cameras and RGB-D cameras have limited depth range and produce poor depth estimates in large-scale outdoor environments.  Second, many egomotion estimation techniques, like visual odometry or visual SLAM, are particularly sensitive to rotation estimates as small early errors have a large influence on final pose estimates. Finally, the constrained nature of rotations presents several difficulties for optimization algorithms. Indeed, if rotations are known, the general problem of pose graph relaxation becomes a linear least squares problem that can be solved with no initial guess for translations \cite{Carlone2015-ud}. 

\subsection{Probabilistic Regression}

In one dimension, given an input $x$, with a target output $y_t$, we desire a probabilistic estimate

\begin{equation}
\Mean{y}, \sigma^2,
\end{equation}
where $\sigma^2$ captures some notion of model uncertainty (owing to the central limit theorem, we will often make the assumption of Gaussian likelihood).

\subsubsection{HydraNet}
One possible way to obtain $\Mean{y}$ is to train a deep neural network, $g(x)$. To endow this network with uncertainty, we present a network structure we call HydraNet (see \Cref{fig:kitti_flow_hydranet}). HydraNet is composed of a large, main `body' with multiple heads that each output a prediction, $g_i(x)$. To compute  $\Mean{y}$, we can simply take the arithmetic mean of the outputs,
\begin{equation}
\label{eq:hn_1d_mean}
\Mean{y} = \frac{1}{H}\sum_{i=1}^{H}g_i(x).
\end{equation} 
The head structure, however, provides several key advantages toward the goal of estimating consistent uncertainty. Namely, it allows us to define the overall uncertainty in terms of two sources, \textit{epistemic} ($\sigmah$) and \textit{aleatoric} ($\sigmad$):
\begin{equation}
\label{eq:1d_hydranet_uncertainty}
\sigma^2 = \sigma^2_{\text{e}} + \sigma^2_{\text{a}}.
\end{equation} 

The former, $\sigmah$, is also sometimes referred to as model uncertainty; it is a measure of how close a particular test sample is to known training samples. The latter, $\sigmad$, is inherent to the observation of the target itself.  Even if the model can localize a test sample exactly in some salient input space, the aleatoric uncertainty will prevent exact regression due to physical processes like sensor noise.

To account for aleatoric uncertainty, we follow prior work \cite{Haarnoja2016-ph,Lakshminarayanan2017} and dedicate one head of the network to regressing a variance directly through a negative log likelihood loss under the assumption of Gaussian likelihood. 

To capture epistemic uncertainty, we train each head with random weight initializations and apply losses independently during training. During test time, we compute a sample covariance over the different outputs. This approach is inspired by the method of the statistical bootstrap \cite{Osband2016}, which predicts population statistics by computing statistics over subsets of a sample chosen with replacement. Unlike \cite{Osband2016}, we do not train each head of the network with a bootstrapped sample, but instead rely on the random initializations of their parameters and the method of dropout to introduce sufficient stochasticity into their outputs. Unlike \cite{Lakshminarayanan2017}, we do not require numerous trained models that can incur high computational cost for complex regression tasks.

%
%


\subsubsection{One-dimensional experiment}

To build intuition for the advantages of HydraNet over other methods of extracting uncertainty (e.g., uncertainty through dropout \cite{Gal2016-ny}), we constructed an experiment similar to that presented in \cite{Osband2016}. We compared HydraNet to four other approaches: (1) direct aleatoric variance regression where the network outputs a second variance parameter that is constrained to be positive, (2) uncertainty through dropout at test time \cite{Gal2016-ny}, (3) bootstrap aggregation (or bagging) of multiple independent models, and (4) HydraNet with no aleatoric uncertainty output.  

\begin{figure*}[h!]
	\centering
	\begin{subfigure}[]{0.32\textwidth}
		\includegraphics[width=\textwidth]{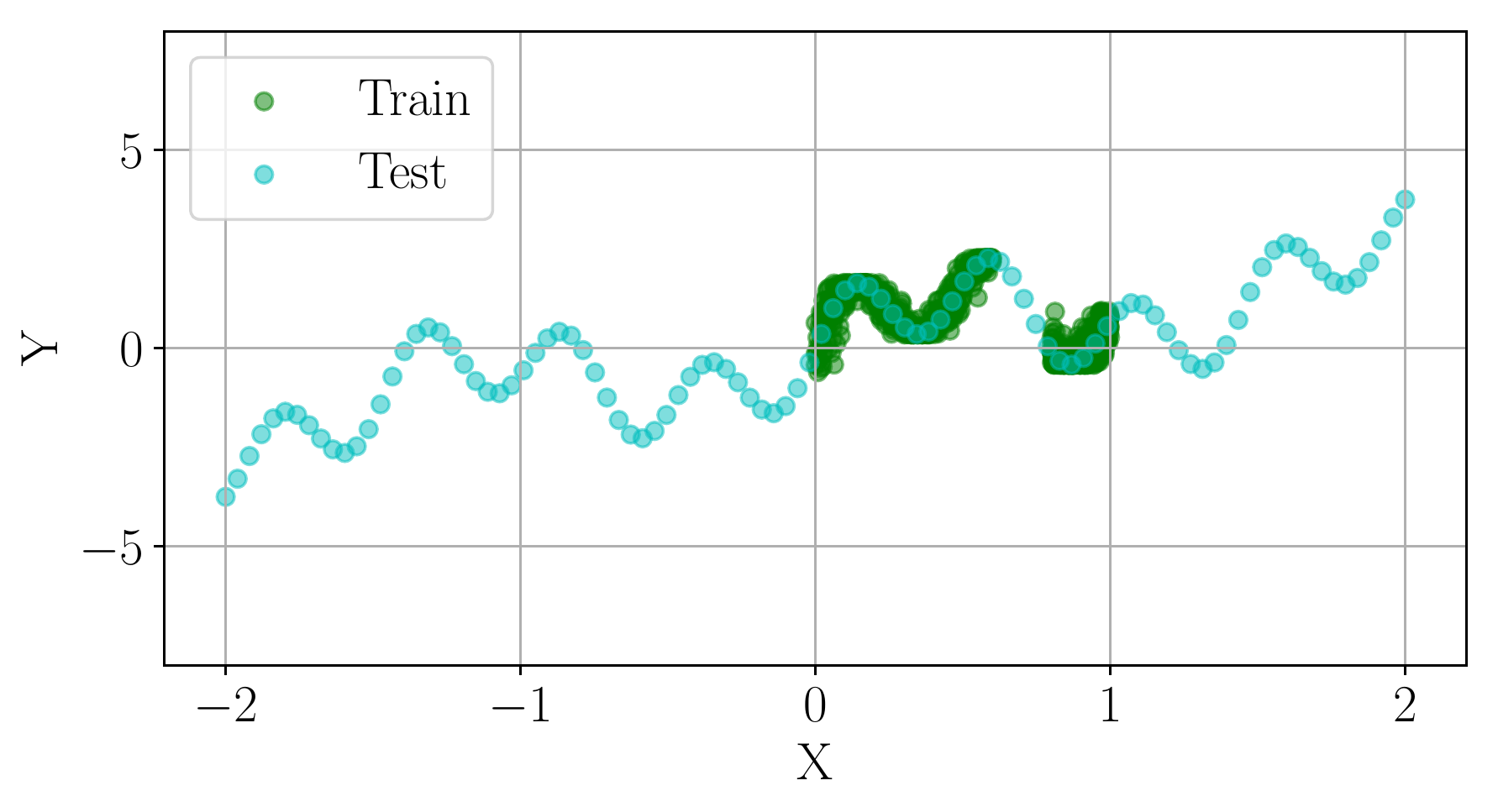}
		\caption{Train and test set data .}
		\label{fig:1D_rawdata}
	\end{subfigure}
	\hfill
	\begin{subfigure}[]{0.32\textwidth}
		\includegraphics[width=\textwidth]{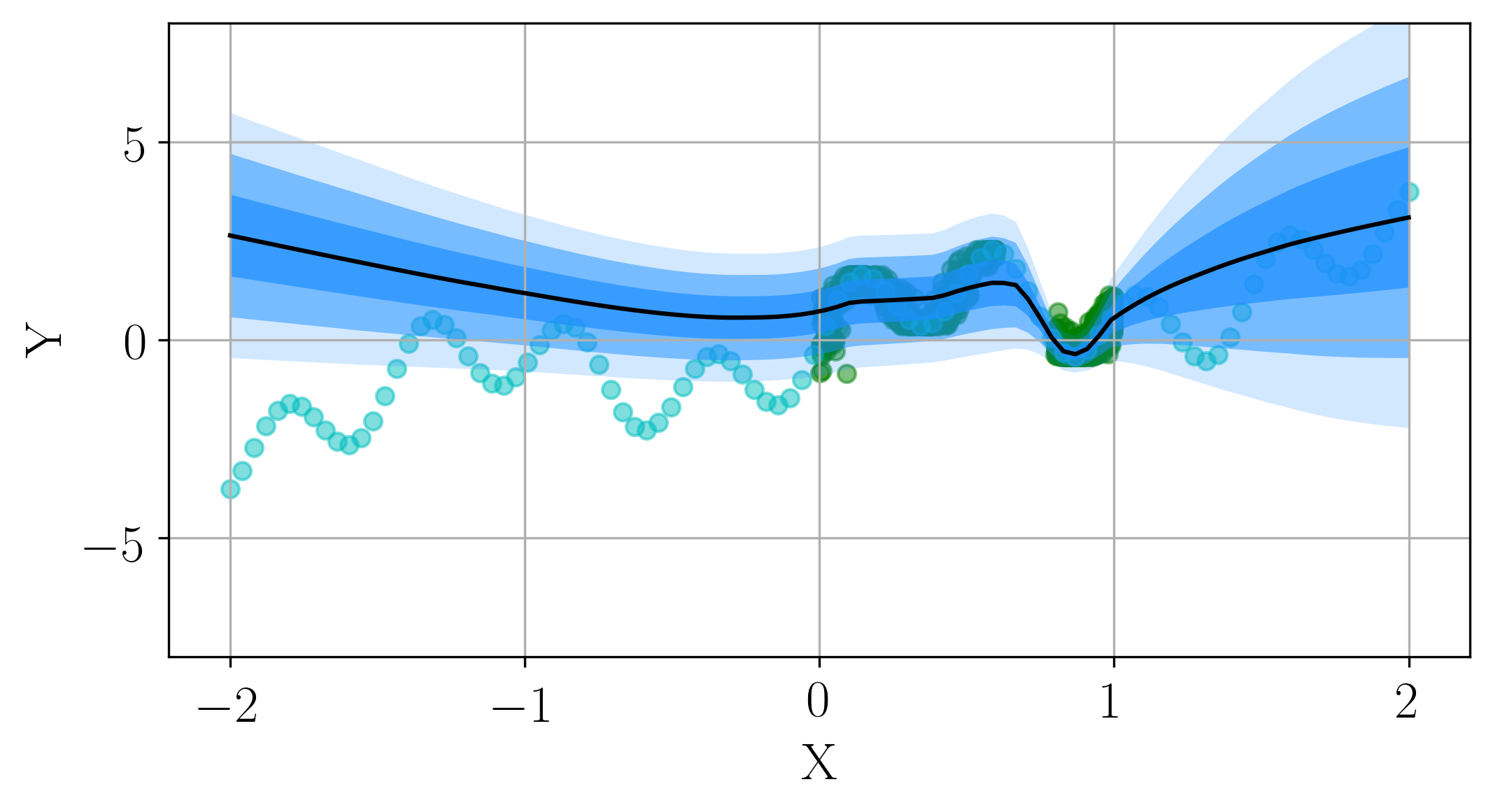}
		\caption{Direct uncertainty regression ($\sigmad$).}
		\label{fig:1D_sigma}
	\end{subfigure}
	\hfill
	\begin{subfigure}[]{0.32\textwidth} 
		\includegraphics[width=\textwidth]{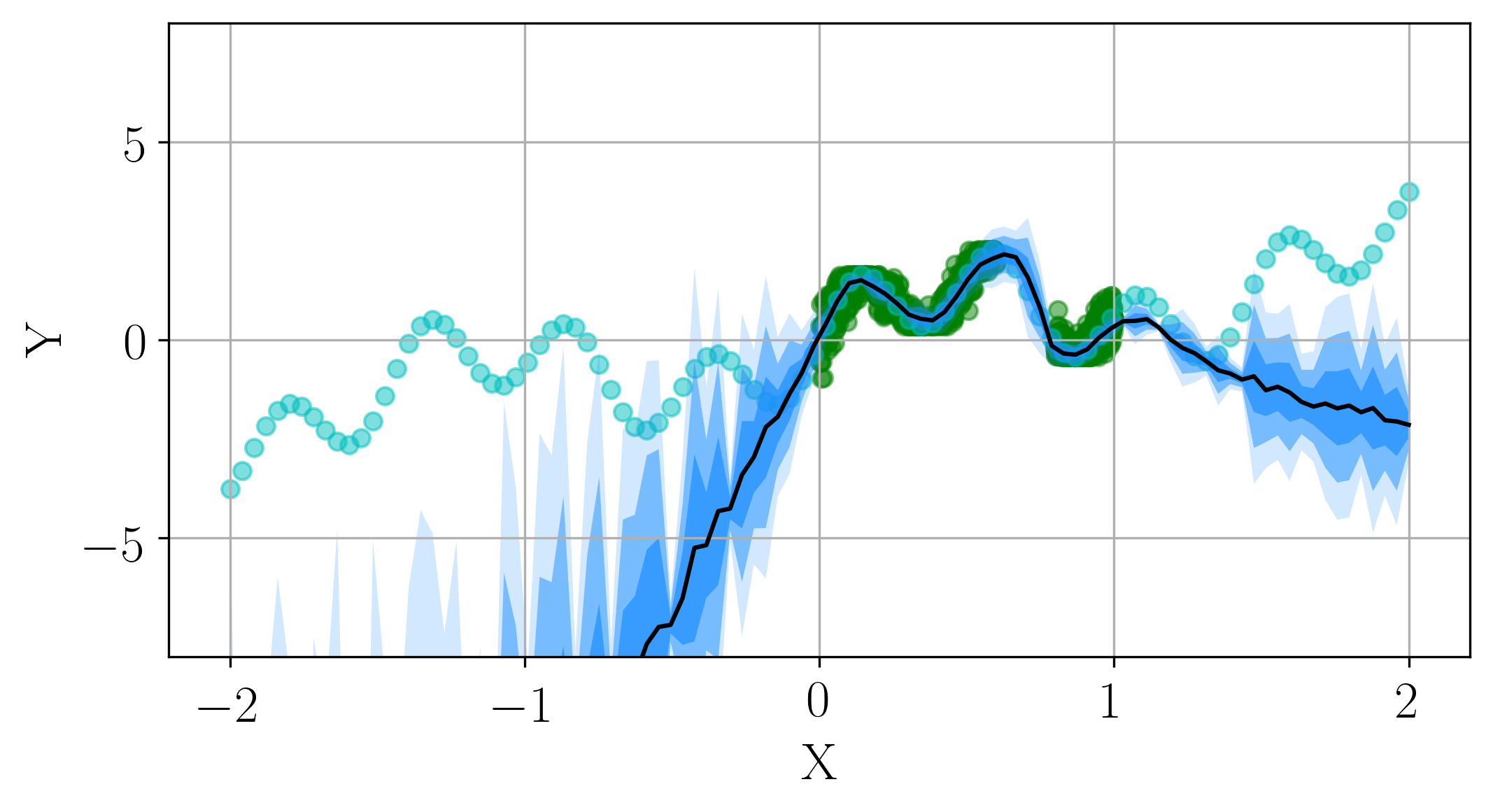}
		\caption{Uncertainty through dropout.}
		\label{fig:1D_dropout}
	\end{subfigure} \\	
	\begin{subfigure}[]{0.32\textwidth}
	\includegraphics[width=\textwidth]{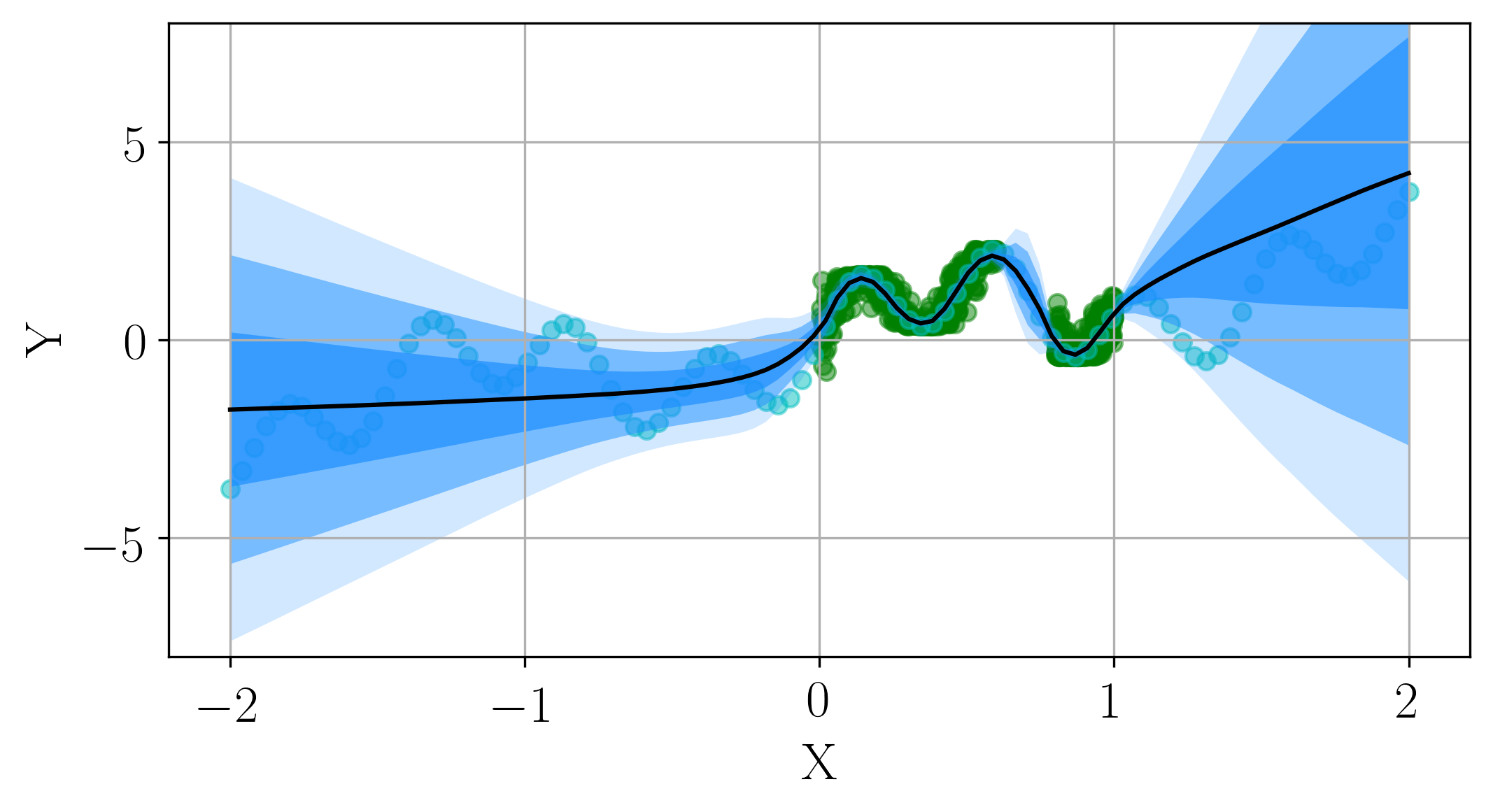}
	\caption{Bootstrap aggregation.}
	\label{fig:1D_ensemble}
	\end{subfigure}
	\hfill
	\begin{subfigure}[]{0.32\textwidth}
		\includegraphics[width=\textwidth]{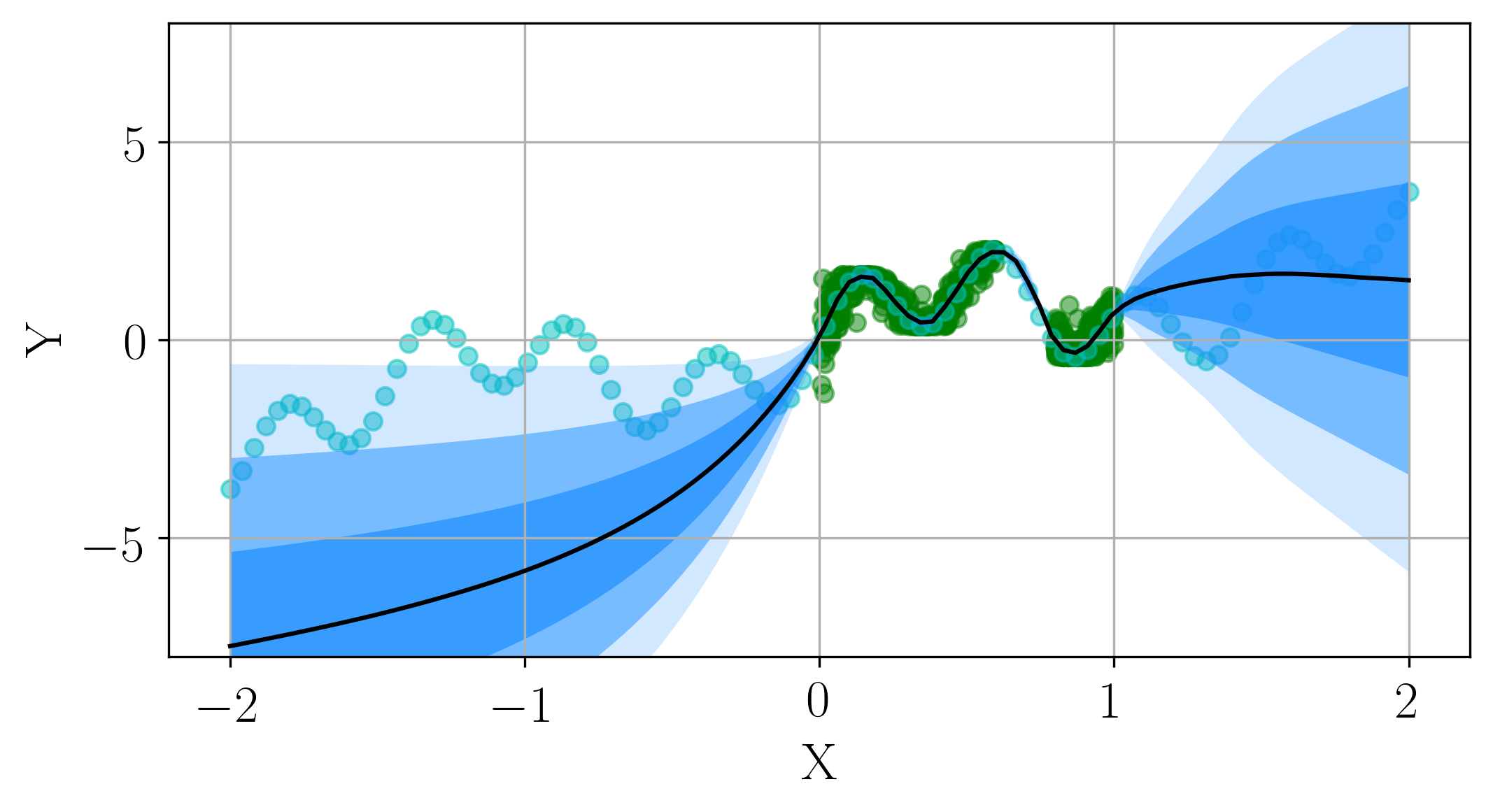}
		\caption{HydraNet (no aleatoric uncertainty).}
		\label{fig:1D_hydra}
	\end{subfigure}
	\hfill
	\begin{subfigure}[]{0.32\textwidth} 
		\includegraphics[width=\textwidth]{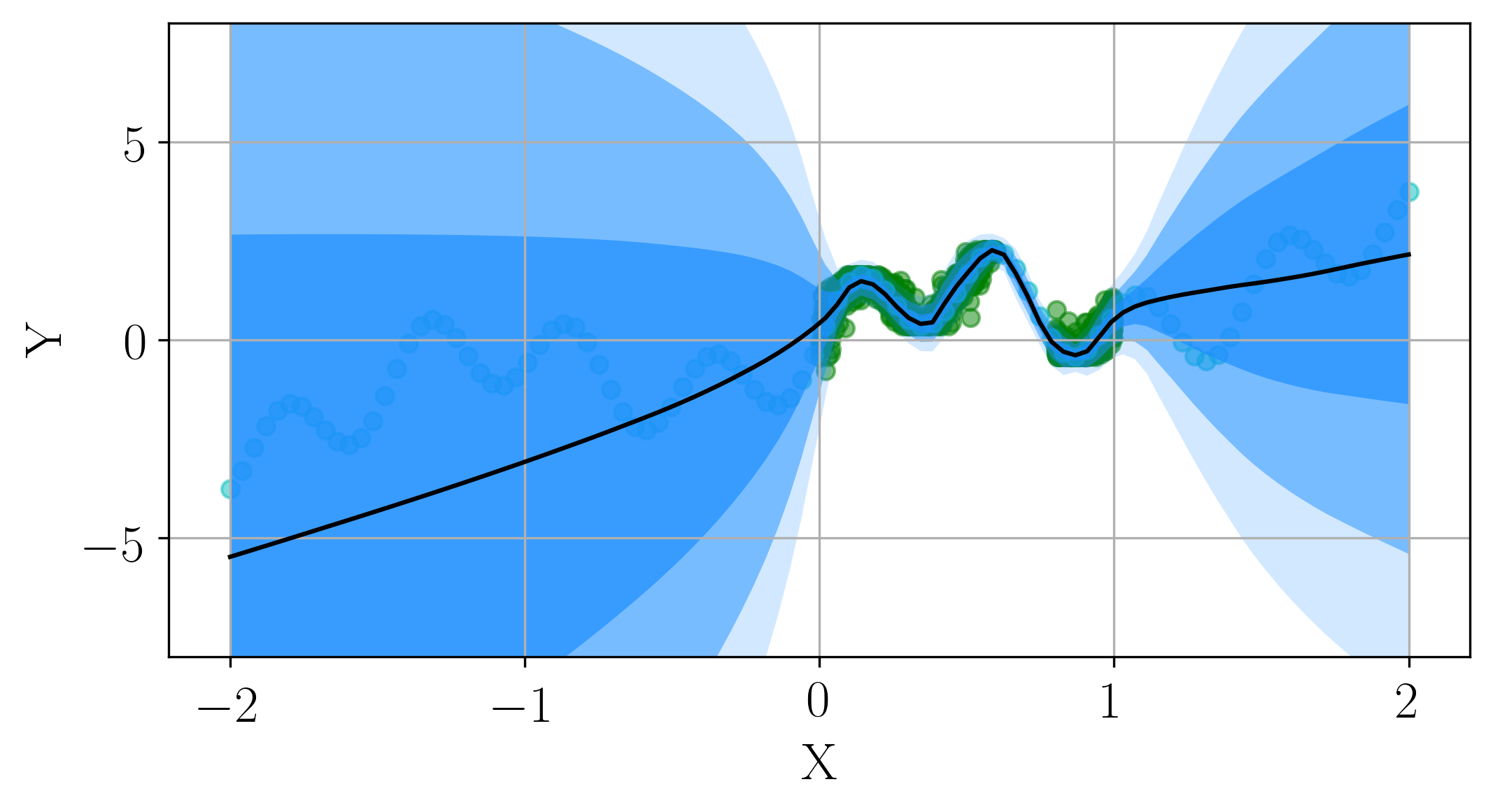}
		\caption{HydraNet.}
		\label{fig:1D_hydrasigma}
	\end{subfigure} \\
		\caption{A comparison of different ways to extract uncertainty from deep networks. Each shade of blue represents one standard deviation $\sigma$ produced by the model.}
		\label{fig:1d_uncertainty}
\end{figure*}

For each method, we trained a four-layer fully-connected network to regress the output of a one-dimensional function: 
\begin{equation}
y_i = x_i + \sin{\left(4( x_i + \omega)\right)} + \sin{\left(13(x_i + \omega)\right)} + \omega,
\end{equation}
where $w \sim \mathcal{N}(\mu=0, \sigma^2=3^2)$. Our training set consisted of 1000 samples randomly drawn from $x \in \left[0.0,0.6\right]\bigcup \left[0.8,1.0\right]$, while the test set consisted of 100 samples uniformly drawn from $x \in \left[ -2, 2\right]$. The function and the train/test samples are shown in \Cref{fig:1D_rawdata}. 




\begin{figure}
	\centering
	\includegraphics[width=0.48\textwidth]{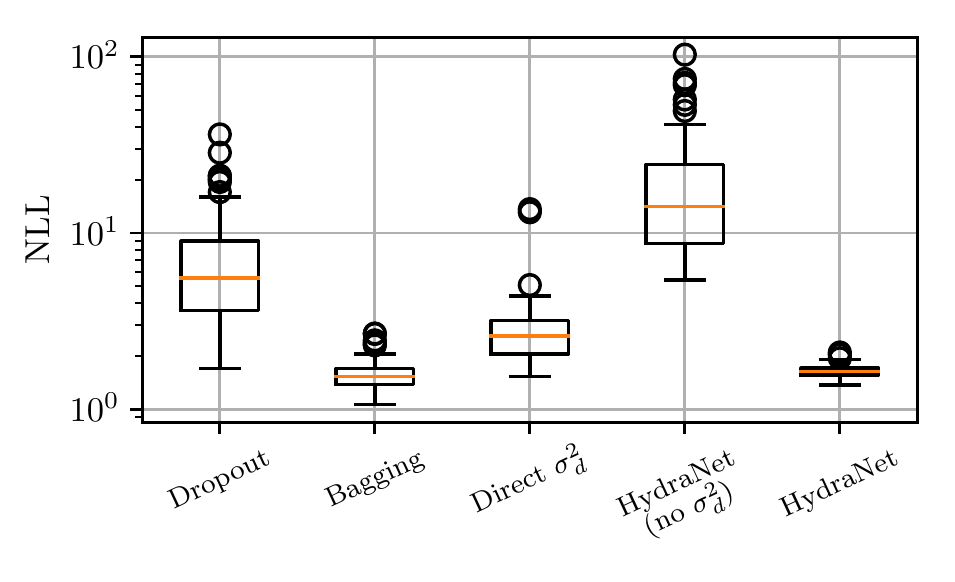}
	\vspace{-1em}
	\caption{Negative log likelihood statistics of 100 repetitions of five neural-network-based uncertainty estimators. HydraNet performs similarly to bagging. }
	\label{fig:noise-free-boxplot}
\end{figure}

The direct aleatoric uncertainty regression and HydraNet methods were trained using a negative log likelihood loss under the assumption of Gaussian likelihood, while the other methods were trained to minimize mean squared error.
We repeated training 100 times, and recorded the test-time negative log likelihood for each method at each repetition.  We summarize the results in \Cref{fig:noise-free-boxplot}.   \Cref{fig:1d_uncertainty} presents representative samples from the 100 repetitions for each method.  Typically, direct uncertainty regression and dropout are overconfident in the out-of-distribution regions. We replicated the findings of \cite{Osband2016} who find that uncertainty with dropout does not vary smoothly and can collapse outside of the training distribution. HydraNet combined with direct aleatoric uncertainty learning, however, produced similar excellent likelihoods to bootstrap aggregation without requiring multiple models.

\subsection{Deep Probabilistic $\LieGroupSO{3}$ Regression}

In order to extend the ideas of HydraNet to the matrix Lie group SO(3), we
consider different ways to regress and combine several estimates of rotation.s
Given a network, $g(\cdot)$, and an input $\mathcal{I}$, we consider how to extend the ideas of HydraNet to process several outputs, $g_i(\mathcal{I})$, and combine them into an estimate of a `mean' rotation, $\Mean{\Matrix{R}}$, and an associated $3\times3$ covariance matrix, $\Matrix{\Sigma}$.  To produce estimates of rotation for a given HydraNet head, we consider two options. First if $g(\mathcal{I}) \in \Real^3$, then we can use the matrix exponential to produce a rotation matrix,
\begin{equation}
	\Matrix{R} = \MatExp{g(\mathcal{I})}.
\end{equation}
Since the capitalized exponential map $\MatExp{\cdot}$ is surjective \cite{Barfoot2017-ri,Sola2018-kg}, this approach can parametrize any valid rotation matrix. Alternatively, if $g(\mathcal{I}) \in \Real^4$, we can normalize it to produce a unit quaternion that resides on $S^3$,
\begin{equation}
\label{eq:quat_output}
	\Vector{q} = \frac{g(\mathcal{I})}{\Norm{g(\mathcal{I})}}.
\end{equation}

Unit quaternions are a double cover of $\LieGroupSO{3}$, and can represent any rotation. We choose to use this latter parametrization because of its simple analytic mean expression that we describe below.

\subsubsection{Rotation Averaging}

To produce a mean of several $\LieGroupSO{3}$ elements (i.e., to evaluate \Cref{eq:hn_1d_mean} for rotations), we turn to the field of rotation averaging \cite{Hartley2013-rc}. Given several estimates of a rotation, we define the mean as the rotation which minimizes some squared metric defined over the group\footnote{Although this is a natural formulation for the rotation mean, it is possible to define other means in terms of absolute errors - see \cite{Hartley2013-rc}.},

\begin{equation}
\label{eq:rot_mean}
\Matrix{\Mean{R}} = \ArgMin{\Matrix{R} \in \LieGroupSO{3}} \sum_{i=1}^n \RotDist{}(\Matrix{R}_i, \Matrix{R})^2.
\end{equation}


There are three common choices for a bijective metric \cite{Hartley2013-rc,Carlone2015-ud} on $\LieGroupSO{3}$. The angular, chordal and quaternionic:
\begin{align}
\RotDist{ang}(\Matrix{R}_a, \Matrix{R}_b) &= \Norm{\MatLog{\Matrix{R}_a \Matrix{R}_b^T}}_2, \\
 \RotDist{chord}(\Matrix{R}_a, \Matrix{R}_b) &= \Norm{\Matrix{R}_a - \Matrix{R}_b}_\mathrm{F}, \\
 \RotDist{quat}(\quat_a, \quat_b) &= \min\left( \Norm{\quat_a - \quat_b}_2, \Norm{\quat_a + \quat_b}_2 \right),
\end{align}
where $\MatLog{\cdot}$, represents the capitalized matrix logarithm \cite{Sola2018-kg}, and $\Norm{\cdot}_F$ the Frobenius norm. In the context of \Cref{eq:rot_mean}, using the angular metric leads to the \textit{Karcher mean}, which requires an iterative solver and has no known analytic expression. Applying the chordal metric leads to an analytic expression for the average but requires the use of Singular Value Decomposition. Using the quaternionic metric, however, leads to a simple, analytic expression for the rotation average as the normalized arithmetic mean of a set of unit quaternions \cite{Hartley2013-rc},
\begin{equation}
\label{eq:quat_mean}
\Mean{\quat} = \ArgMin{\Matrix{R}(\quat) \in \LieGroupSO{3}} \sum_{i=1}^H \RotDist{quat}(\Matrix{q}_i, \Matrix{q})^2 = \frac{\sum_{i=1}^H \quat_i}{\Norm{\sum_{i=1}^H \quat_i}}.
\end{equation}

This expression is simple to evaluate numerically, and if necessary, can be easily differentiated with respect to its constituent parts. For these reasons, we opt to construct our $\LieGroupSO{3}$ HydraNet using unit quaternion outputs, and evaluate the rotation average using the quaternionic metric.

\subsubsection{SO(3) Uncertainty}
There are several ways to approach uncertainty on $\LieGroupSO{3}$. One method \cite{Carlone2015-aq} is to define a probability density directly on the group via the isotropic von Mises-Fisher density. This approach has two downsides: (1) it is isotropic and cannot account for dominant  degrees of freedom (e.g., vehicle yaw during driving), and (2) estimating the concentration parameter requires approximations or iterative solvers \cite{Hornik2014-jw}.

Instead, we opt to parametrize uncertainty over $\LieGroupSO{3}$ by injecting uncertainty onto the manifold \cite{Forster2015-af,Barfoot2014-ac,Barfoot2017-ri} from a local tangent space about some mean element, $\Mean{\quat}$,
\begin{equation}
\label{eq:quat_inject_density}
\quat = \MatExp{\Vector{\epsilon}} \otimes \Mean{\quat}, ~~ \Vector{\epsilon} \sim \NormalDistribution{\Vector{0}}{\Matrix{\Sigma}},
\end{equation}
 \noindent where $\otimes$ represents quaternion multiplication. In this formulation, $\Matrix{\Sigma}$ provides a $3\times 3$ covariance matrix that can express uncertainty in different directions. Further, given a mean rotation, $\Mean{\quat}$, and samples, $\quat_i$, we use the logarithmic map to compute a sample covariance matrix,\begin{equation}
\label{eq:head_covariance}
\Covh = \frac{1}{H-1} \sum_{i=1}^H  \Vector{\phi}_i \Vector{\phi}_i^T, ~~ \Vector{\phi}_i = \MatLog{\quat_i \otimes \Mean{\quat}^{-1}}.
\end{equation}

\subsection{Loss Function}

As with one-dimensional HydraNet, we train a direct regression of covariance through a parametrization of positive semi-definite matrices using a Cholesky decomposition\footnote{Note that in all the experiments presented in this paper, we omit the off-diagonal components of this covariance and only learn a diagonal matrix with non-negative components.} \cite{Hu2015-uw,Haarnoja2016-ph}). Given the network outputs of a unit quaternion $\quat$, and a positive semi-definite matrix $\Sigma$, we define a loss function as the negative log likelihood of a given rotation under \Cref{eq:quat_inject_density} (see \cite{Forster2015-af}) for a given target rotation, $\quat_t$, as
\begin{equation}
\label{eq:nll_loss}
\mathcal{L}_\mathrm{NLL}(\quat, \quat_t, \Covd) = \frac{1}{2}\Vector{\phi}^T \Covd^{-1}\Vector{\phi} + \frac{1}{2} \log{\Determinant{\Covd}},
\end{equation}
\noindent where  $\Vector{\phi} = \MatLog{\quat \otimes {\quat_t}^{-1}}$. Combining the sample covariance, with the learned covariance, we extend \Cref{eq:1d_hydranet_uncertainty} to
\begin{equation}
\Matrix{\Sigma}_t = \Covh + \Covd.
\end{equation}
This covariance estimate is designed to grow for out-of-training-distribution errors (and account for \textit{domain shift} \cite{Lakshminarayanan2017}) while still accounting for uncertainty within the training set. We note that unlike Bayesian methods, we do not interpret each head as a \textit{sample} from a posterior distribution\footnote{Notably, this means we do not scale our direct uncertainty when averaging as $\frac{1}{H}\Covd$.}. Indeed, we note that in our 1D experiments, the heads have very small variance within the training distribution. The multi-headed structure and rotating averaging serves simply as a way to model epistemic uncertainty when the model encounters inputs that differ from those seen during training. We summarize our training and test procedures in \Cref{alg:train_hydranet} and \Cref{alg:test_hydranet} respectively.

\begin{algorithm}
  \caption{Supervised training for $\LieGroupSO{3}$ regression}
   \label{alg:train_hydranet}
   \begin{spacing}{1.1}
  \begin{algorithmic}[1]
    \Require{Training data $\mathcal{T}$, training targets $\quat_t$, untrained model $g_\theta(\cdot)$ with parameters $\theta$ and $H+1$ heads}
    \Ensure{Probabilistic regression model $g_\theta(\cdot)$}
    \Function{TrainHydraNet}{$\mathcal{T}$} 
    \For{each mini-batch $\mathcal{T}_i$}
    
    \State Output $\Covd$ \Comment{\emph{1st head, Chol. decom.}}
    \For{heads $2...(H+1)$ in $g$}
    \State Output  $\quat_h$ \Comment{\emph{\Cref{eq:quat_output}}}
    \State Evaluate NLL loss \Comment{\emph{\Cref{eq:nll_loss}}}
    \EndFor{end}
     \State Backprop, update $\theta$
     \EndFor{end}
    \State \Return{$g(\cdot)$}
    \EndFunction
  \end{algorithmic}
  \end{spacing}
\end{algorithm}

\begin{algorithm}
  \caption{Testing of $\LieGroupSO{3}$ regression}
   \label{alg:test_hydranet}
   \begin{spacing}{1.1}
  \begin{algorithmic}[1]
    \Require{Test sample $\mathcal{I}_j$, trained model $g_\theta(\cdot)$}
    \Ensure{Test prediction $\quat$, covariance $\Matrix{\Sigma}_t \succcurlyeq 0 $}
    \Function{TestHydraNet}{$\mathcal{I}_j$, $g_\theta(\cdot)$} 
     \State Output $\Covd$ \Comment{\emph{1st head, Chol. decom.}}
    \For{heads $2...(H+1)$ in $g$}
    \State Output  $\quat_h$ \Comment{\emph{\Cref{eq:quat_output}}}
    \EndFor{end}
    \State Compute $\Mean{\quat}$ \Comment{\emph{\Cref{eq:quat_mean}}}
    \State Compute $\Covh$  \Comment{\emph{\Cref{eq:head_covariance}}}
	\State \Return{$\Mean{\quat}$, $\Covh + \Covd$}
    \EndFunction
  \end{algorithmic}
  \end{spacing}
\end{algorithm}
\section{Experiments}
\subsection{Uncertainty Evaluation: Synthetic Data}

\begin{figure}
	\centering 
	\includegraphics[width=0.48\textwidth]{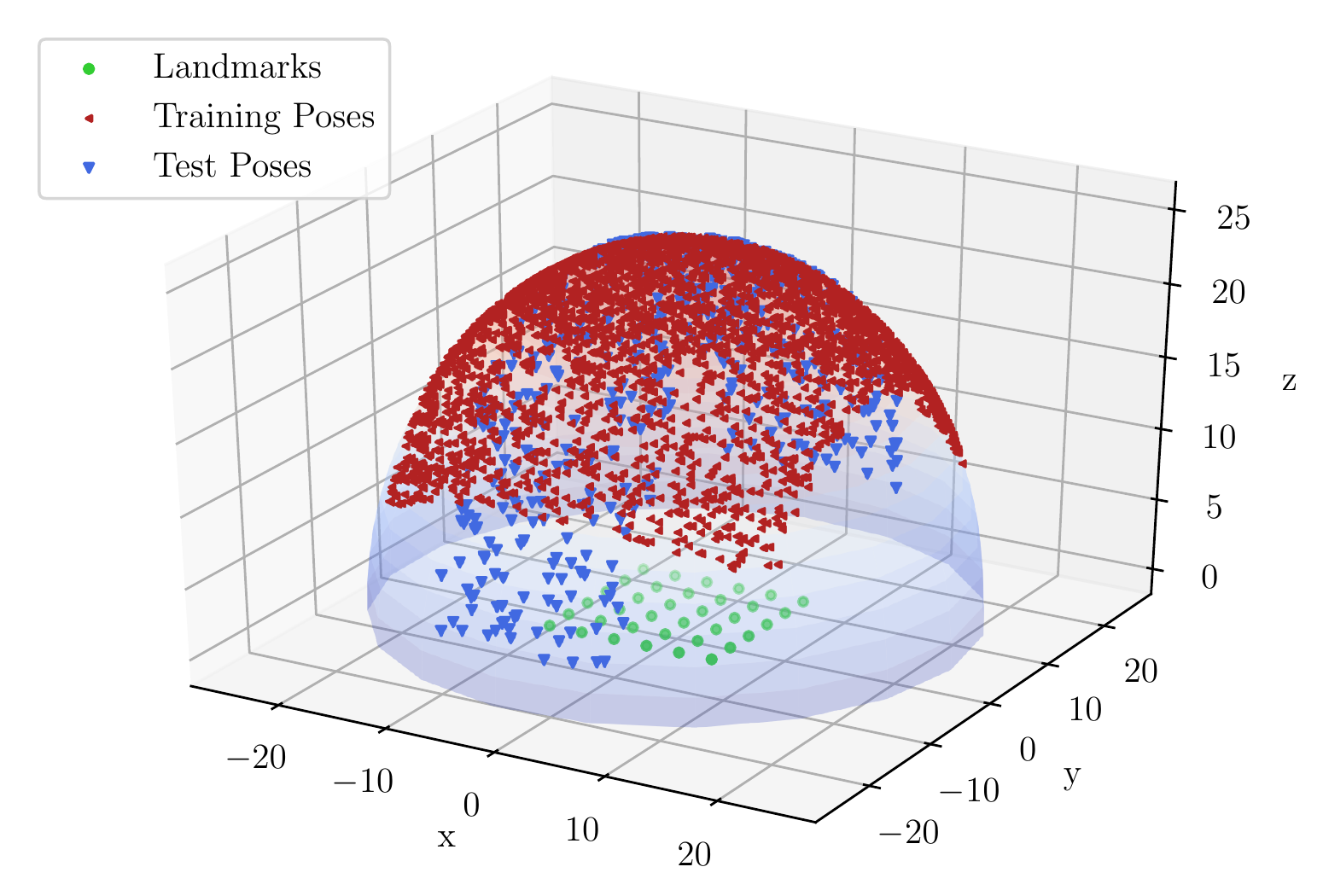}
	\caption{Synthetic world used to illustrate our method. A monocular camera observes a 6 $\times$ 6 grid of point landmarks from poses sampled on a semi-sphere. The test set includes poses that are outside the training distribution.}
	\label{fig:synthetic_world}
	\vspace{-1.5em}
\end{figure}

Before we embarked on training with real data, we analyzed our proposed HydraNet structure on a synthetic world. Our goal was to produce probabilistic estimates of camera orientation based on noisy pixel coordinates of a set of fixed point landmarks. To accomplish this, we simulated a monocular camera observing a planar grid of evenly spaced (see \Cref{fig:synthetic_world}) landmarks from a hemisphere surrounding the grid. We aligned the monocular camera's optical axis with the centre of the hemisphere so that all landmarks were visible in every camera pose. At each pose, we computed noisy pixel locations of the projection of every landmark, and stacked these 2D locations as an input vector. We generated 15000 training samples with poses that were randomly sampled from the hemisphere in the polar angle range of $[-60, 60]$ degrees. For testing, we sampled 500 poses in the range of $[-80, 80]$ degrees, purposely widening the range to include orientations that were not part of training. 

\begin{figure}
	\centering
	\includegraphics[width=0.49\textwidth]{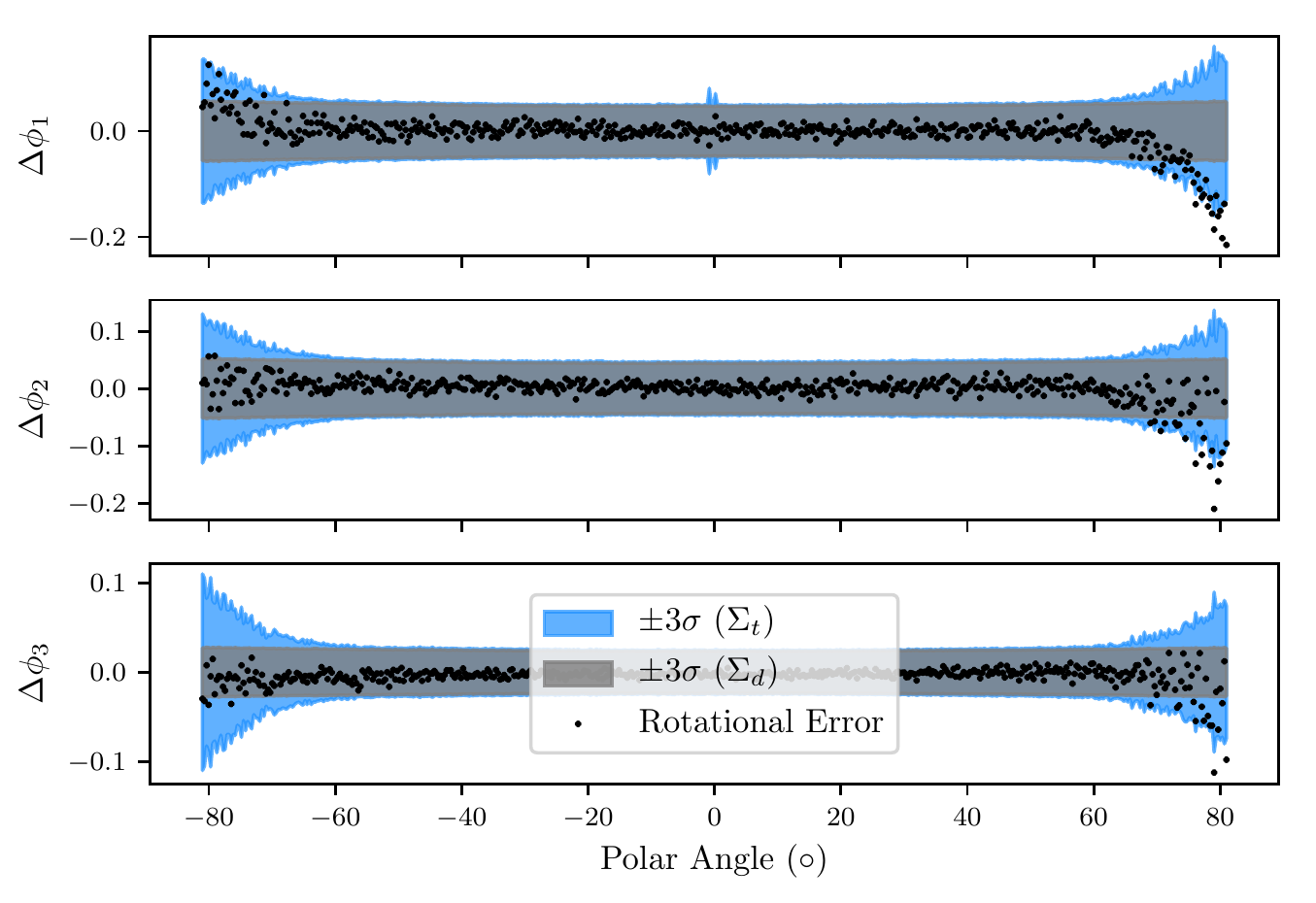}
	\caption{Rotation estimation errors for a deep network trained using our HydraNet approach on synthetic data (noisy pixel locations of 36 landmarks). We note that outside of the training distribution, our epistemic uncertainty ($\Covh$) grows, as expected.}
	\label{fig:sim_errors}
\end{figure}

To regress the camera orientation, we constructed a five layer residual network and attached 26 heads (25 + 1 for direct uncertainty learning) to regress a probabilistic estimate of $\quat_{c,w}$, the orientation of the camera with respect to the world frame. 

\Cref{fig:sim_errors} plots rotational errors $\Vector{\phi} = \MatLog{\quat \otimes \quat_t^{-1}}$  along with 3 sigma bounds based on both the total covariance, $\Matrix{\Sigma}_t$, and the direct covariance $\Covd$. The final regression estimates have consistent uncertainty, composed of a static aleatoric uncertainty  and an epistemic uncertainty (\Cref{eq:head_covariance}) that grows when the test samples come from unfamiliar input data.

\subsection{Absolute Orientation: 7-Scenes}
Next, we used HydraNet to regress absolute orientations from RGB images from the 7-Scenes dataset \cite{Glocker7scenes}. Our goal was to achieve similar errors to other regression techniques \cite{Kendall2017-ud} but augment them with consistent covariance estimates. For this experiment, we used \texttt{resnet34} \cite{he2016deep} (pre-trained on the ImageNet dataset) for the body of HydraNet and attached 25 HydraNet heads, each consisting of two fully connected layers. We cropped and resized all RGB images to match the expected ImageNet size and omitted the depth channel.

\Cref{tab:7scenes_stats} presents the mean angular errors and negative log likelihoods achieved by our method. The HydraNet-based network produces similar angular errors to other regression methods \cite{Kendall2017-ud} but with additional benefit of consistent three-degree-of-freedom uncertainty. Note that we spent little time optimizing the network itself, and note that state-of-the art errors can be achieved using more sophisticated pixel-based losses \cite{Brachmann2018-us}. However, the general HydraNet structure and loss can be used whenever a probabilistic rotation output is required. Further, our results show that our covariance formulation can be used for `large' rotation elements, where techniques (e.g., \cite{2018_Peretroukhin_Deep}) that assume `small' corrections may fail.

\begin{table}[]
	\caption{HydraNet regression results for the 7scenes dataset compared to results reported in \cite{Kendall2017-ud}. We report mean angular errors and the negative log likelihood (lower is better).}
	\begin{threeparttable}
	\begin{tabular}{cccccc}
		\toprule
		& \multicolumn{2}{c}{\textbf{Error (deg)}} & \multicolumn{2}{c}{\textbf{NLL}} \\ \cmidrule{2-5} 
		\textbf{Scene} & HydraNet & PoseNet & HydraNet & PoseNet \\ \midrule
		Chess & 6.3 & 4.5 & -6.0 & --- \\
		Fire & 14.9 & 11.3 & -3.6 & --- \\
		Heads & 14.3 & 13.0 & -3.9 & --- \\
		Office & 8.6 & 5.6 & -5.4 & --- \\
		Pumpkin & 9.0 & 4.8 & -5.0 & --- \\
		Kitchen & 8.8 & 5.4 & -5.0 & --- \\
		Stairs & 11.8 & 12.4 & -4.7 & --- \\ \bottomrule
	\end{tabular}
\end{threeparttable}
\label{tab:7scenes_stats}
\end{table}

\begin{figure}
	\centering
	\includegraphics[width=0.49\textwidth]{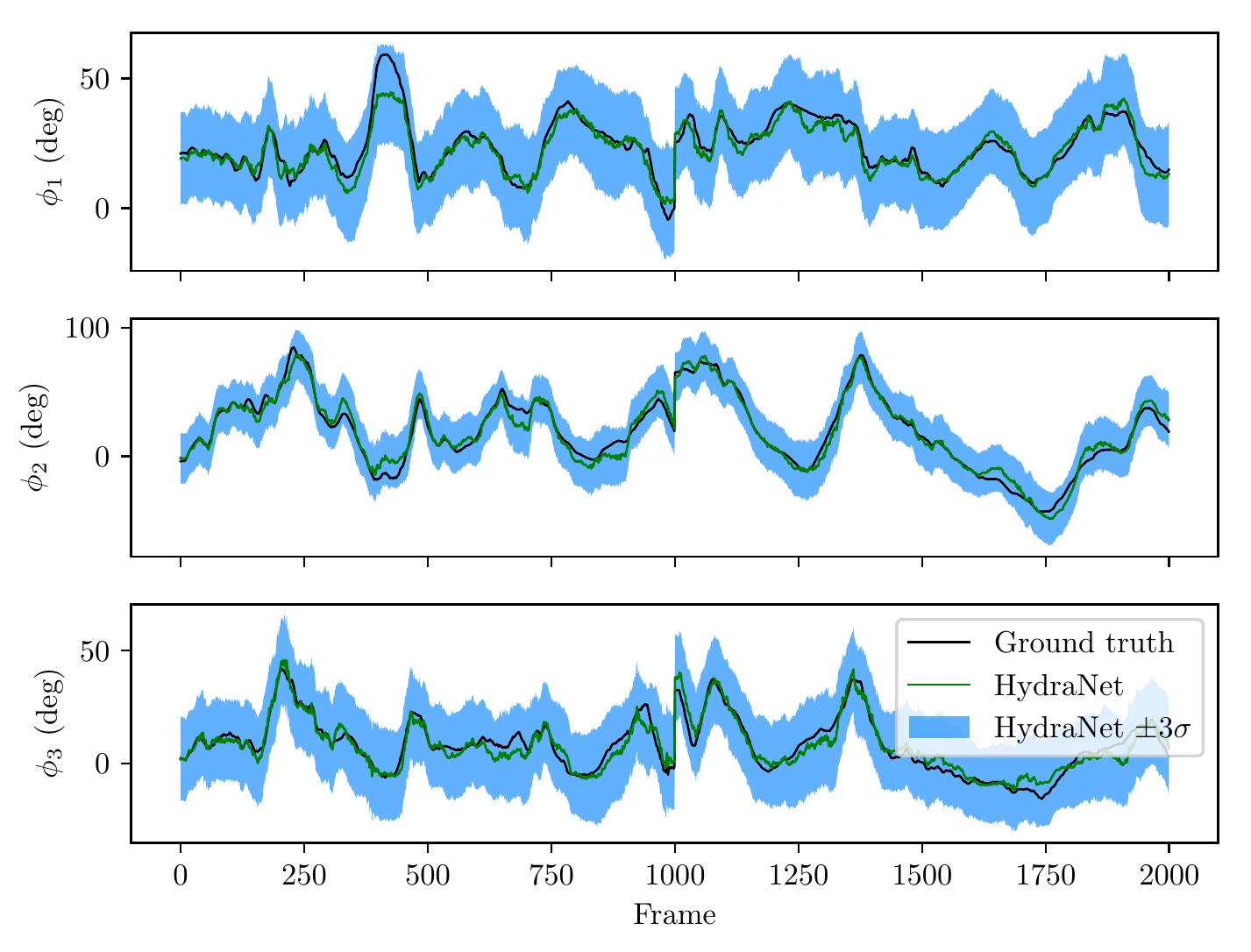}
	\caption{Orientation regression results for the 7scenes \textit{chess} test set. Our HydraNet structure paired with a \texttt{resnet-34} results mean errors of 6.3 degrees, with consistent uncertainty. We detail results for all seven scenes in \Cref{tab:7scenes_stats}.}
	\label{fig:7scenes_chess_absolute}
	\vspace{-1.5em}
\end{figure}


\subsection{Relative Rotation: KITTI Visual Odometry}
  
\begin{table*}[]
	\caption{Results of fusing HydraNet relative rotation regression with classical stereo visual odometry.}
	\begin{threeparttable}
	\begin{tabular}{cccccc}
		\toprule
		& \textbf{} & \multicolumn{2}{c}{\textbf{m-ATE}} & \multicolumn{2}{c}{\textbf{Mean Segment Errors}} \\ \cmidrule{2-6} 
		\textbf{Sequence (Length)} & Estimator & Translation (m) & Rotation ($^{\circ}$) & Translation (\%) & Rotation ($^{\circ}$/100m) \\ \midrule
		\multirow{6}{*}{\texttt{00} (3.7 km)}  & DeepVO \cite{Wang:2017} & --- & --- & --- & --- \\
		& SfMLearner \cite{Zhou:2017} & --- & --- & 65.27 & 6.23 \\
		& UnDeepVO \cite{Li:2017} & --- & --- & 4.14 & 1.92 \\
		& \texttt{viso2-s} & 27.91 & 6.25 & 1.96 & 0.81 \\
		& \texttt{viso2-s} + HydraNet & 9.86 & 2.83 & 1.34 & 0.63 \\
		& Keyframe Direct VO & 12.41 & 2.45 & 1.28 & 0.54 \\ \midrule
		\multirow{6}{*}{\texttt{02} (5.1 km)} & DeepVO & --- & --- & --- & --- \\
		& SfMLearner & --- & --- & 57.59 & 4.09 \\
		& UnDeepVO & --- & --- & 5.58 & 2.44 \\
		& \texttt{viso2-s}& 64.67 & 8.45 & 1.47 & 0.56 \\
		& \texttt{viso2-s} + HydraNet & 50.19 & 6.51 & 1.47 & 0.63 \\
		& Keyframe Direct VO & 16.33 & 3.19 & 1.21 & 0.47 \\ \midrule
		\multirow{6}{*}{\texttt{05} (2.2 km)} & DeepVO & --- & --- & 2.62 & 3.61 \\
		& SfMLearner & --- & --- & 16.76 & 4.06 \\
		& UnDeepVO & --- & --- & 3.40 & 1.50 \\
		& \texttt{viso2-s} & 23.72 & 8.10 & 1.79 & 0.79 \\
		& \texttt{viso2-s} + HydraNet & 9.85 & 3.23 & 1.38 & 0.60 \\
		& Keyframe Direct VO & 5.83 & 2.05 & 0.69 & 0.32 \\ \bottomrule
	\end{tabular}
\label{tab:kitti_fusion_stats}
\end{threeparttable}
\end{table*}

\begin{figure}
	\centering
	\includegraphics[width=0.49\textwidth]{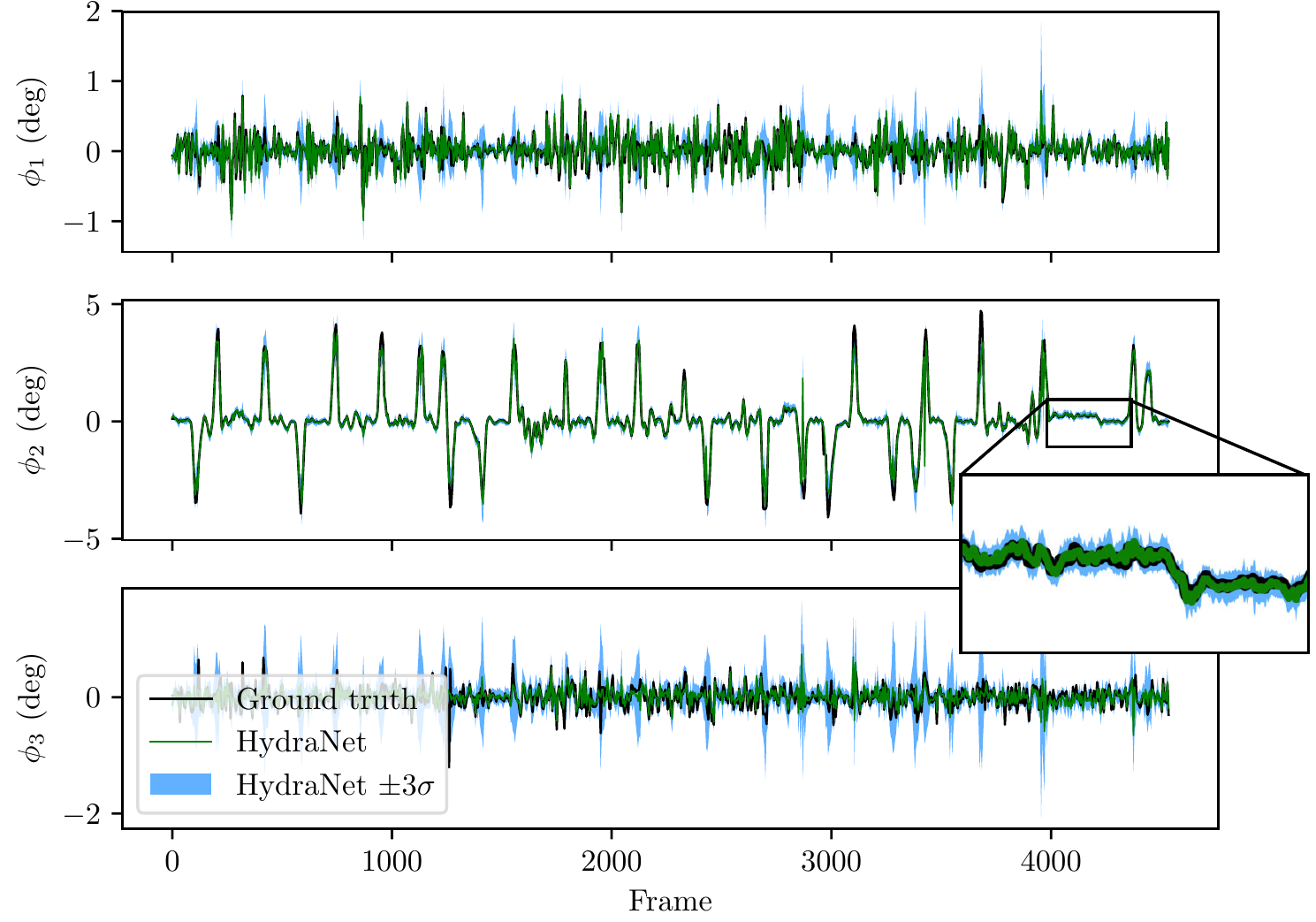}
	\caption{Frame-to-frame rotation regression for KITTI odometry dataset sequence \texttt{00}. Note how the uncertainty increases when the car turns ($\phi_2$ represents the yaw angle). For plotting clarity, we downsample the data from 10Hz to 2Hz. Full statistics can be found in \Cref{tab:kitti_hydranet_stats}.}
	\label{fig:kitti_topdown}
	\vspace{-.5em}
\end{figure}

\begin{figure}
	\centering
	\includegraphics[width=0.49\textwidth]{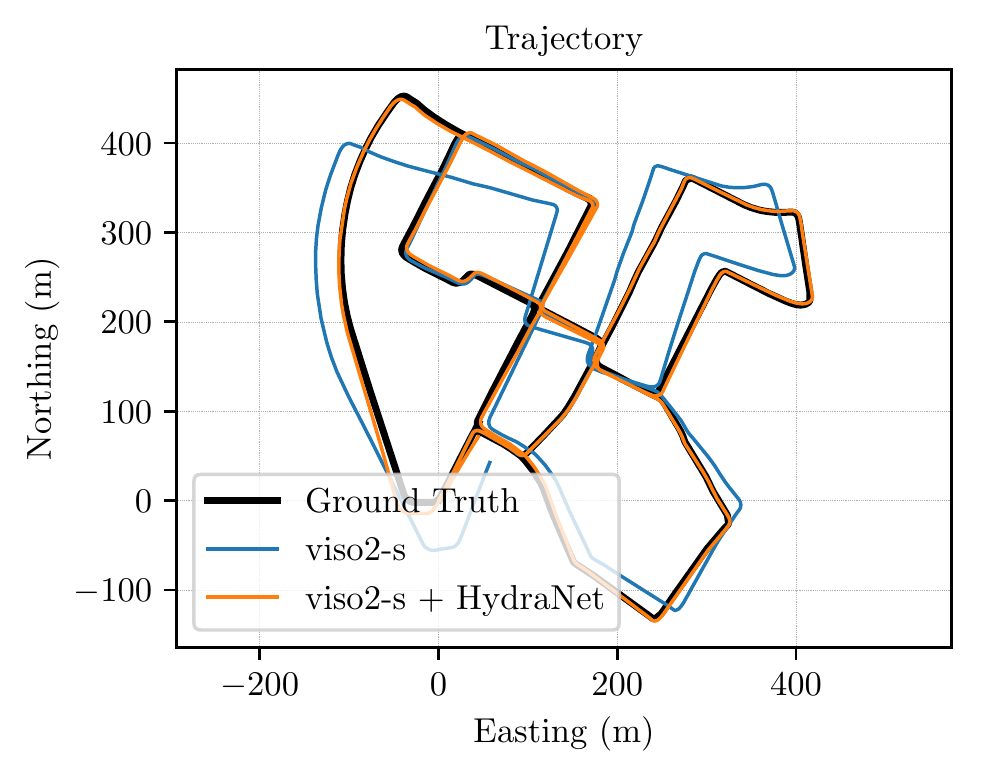}
	\caption{Top-down trajectory of KITTI odometry dataset sequence \texttt{00}.}
	\label{fig:kitti_topdown}
	\vspace{-.5em}
\end{figure}


\begin{table}[]
\centering
\caption{HydraNet regression results for the KITTI odometry dataset. We report mean angular errors and the negative log likelihood (lower is better).}
	\begin{threeparttable}
	\begin{tabular}{ccc} \toprule
		\textbf{Sequence}  & \textbf{Mean Angular Error ($^{\circ}$)} & \textbf{NLL} \\ \midrule
		\texttt{00} & 0.199 & -16.84 \\
		\texttt{02} & 0.138 & -18.44 \\
		\texttt{05} & 0.109 & -19.31 \\
	 \bottomrule
	\end{tabular}
	\end{threeparttable}
	\label{tab:kitti_hydranet_stats}
\end{table}

\begin{figure}
	\centering
	\includegraphics[width=0.49\textwidth]{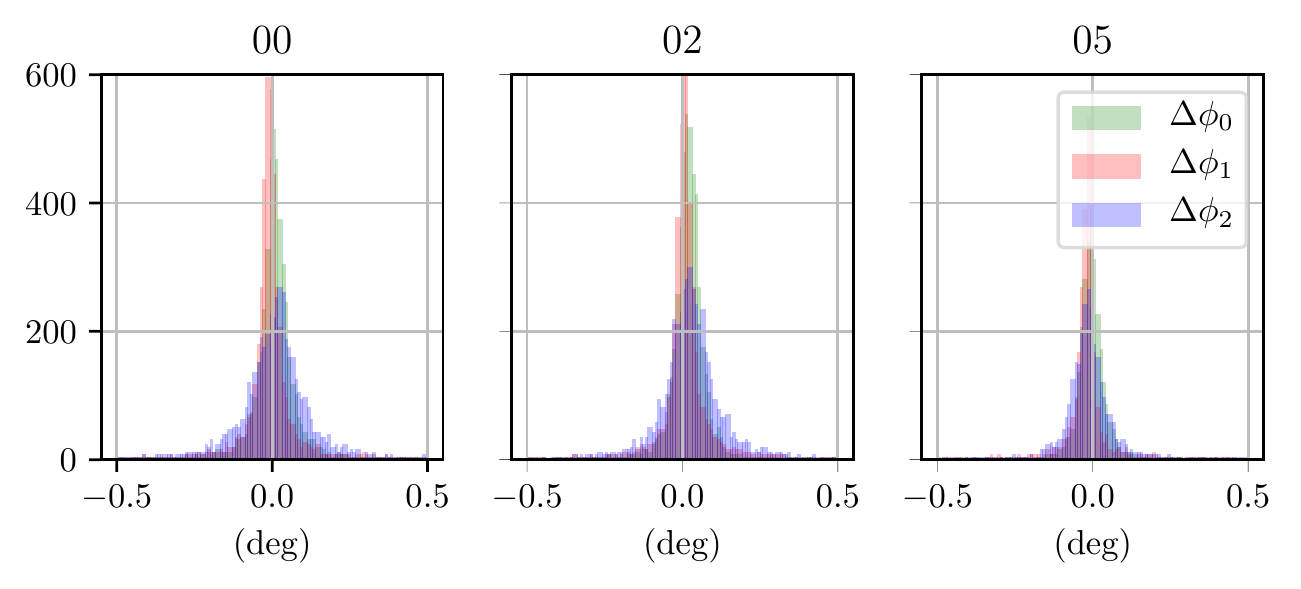}
	\caption{Error histograms for test KITTI sequences \texttt{00}, \texttt{02}, and \texttt{05} on three rotational axes.}
	\label{fig:kitti_hists}
	\vspace{-.5em}
\end{figure}

Finally, to show the benefit of fusing deep probabilistic estimates with classical estimators, we trained a network to estimate relative frame-to-frame rotations on the KITTI visual odometry (VO) benchmark. To regress relative rotations, we use the HydraNet-based network described in \Cref{fig:kitti_flow_hydranet}. For each pair of poses, we process two RGB images (taken from the left RGB camera) into a two channel dense optical flow image using a fast classical algorithm \cite{farneback2003two}. Compared to using raw images, we found that using the optical flow pre-processing greatly improved training robustness and rotation accuracy. Since we use two-channel flow images, the body of the network is not pre-trained and instead contains an eight layer convolutional network. We maintained the same head structure as the 7-Scenes experiment. \Cref{tab:kitti_hydranet_stats} and \Cref{fig:kitti_hists} detail the mean test error and negative log likelihood for KITTI odometry sequences \texttt{00}, \texttt{02} and \texttt{05} (chosen for their complexity and length). For each sequence, we trained the model on the remaining sequences in the benchmark. We found our model produced mean errors of approximately 0.1 degrees on all three test sequences. The covariance produced by HydraNet was consistent, spiking during yawing motions when the largest errors occurred (see \Cref{fig:kitti_topdown}). Despite its consistency, the network covariance was dominated by $\Covd$. We suspect that unlike the synthetic data, $\Covh$ remained small throughout the tests sets due to a more constrained input space (RGB or flow images, compared to pixel locations), but leave a thorough investigation to future work. 
 
\subsubsection{Classical VO}
For the classical visual odometry estimator, we used the open-source \texttt{libviso2} package~\cite{Geiger2011-xe} to detect and track sparse stereo image key-points in a similar manner to \cite{Peretroukhin2018}. In brief, our pipeline modelled stereo re-projection errors, $ \Vector{e}_{l,t_i}$, as zero-mean Gaussians with a known static covariance, $\ImageCovariance$. To generate an initial guess and to reject outliers, we used three point Random Sample Consensus (RANSAC) based on stereo re-projection error.
Finally, we solved for the maximum likelihood transform, $\Transform_{t+1,t}^*$, through a Gauss-Newton minimization of\begin{equation}
  \Transform_{t_{i+1},t_i}^* = \ArgMin{\Transform_{t_{i+1},t_i}\in\text{SE}(3)}\sum_{l=1}^{N_{t_i}} 
  \Transpose{\Vector{e}}_{l} \ImageCovariance^{-1} \Vector{e}_{l}.
\end{equation}
\noindent After convergence, we approximate the frame-to-frame transformation uncertainty as \cite{Barfoot2017-ri}:
\begin{equation}
	\Matrix{\Sigma}_\text{vo} \approx  \left( \sum_{l=1}^{N_t} \Transpose{\Matrix{J}}_{l} \ImageCovariance^{-1} \Matrix{J}_{l} \right)^{-1},
    \label{eq:svo_uncertainty}
\end{equation}
where $\Matrix{J}_{l}$ refers to the Jacobian of each reprojection error.

\subsubsection{Fusion via Graph Relaxation}

To fuse these estimates with classical VO, we used pose graph relaxation. We describe our method briefly and refer the reader to \cite{Barfoot2017-ri} for a more detailed treatment. For every two poses, we defined a loss function based on a contribution from the estimator and from the network, weighed by their respective covariances:
\begin{align}
	\Transform_{1,w}^*, \Transform_{2,w}^* &= \ArgMin{\Transform_{1,w}, \Transform_{2,w}\in\text{SE}(3)}\mathcal{L}(\Estimate{\Transform}_{2,1}, \Estimate{\Rotation}_{2,1}) \\ & = \delta\Vector{\xi}_\text{1,2}^T \Matrix{\Sigma}^{-1}_\text{vo} \delta\Vector{\xi}_\text{1,2} + \delta\Vector{\phi}_\text{1,2}^T \Matrix{\Sigma}^{-1}_{\text{hn}} \delta\Vector{\phi}_\text{1,2} 
\end{align}
where $\delta\Vector{\xi}_\text{1,2} =  \MatLog{\left(\Transform_{2,w} \Transform_{1,w}^{-1} \right)\Estimate{\Transform}_{2,1}^{-1}}$ and
$\delta\Vector{\phi}_\text{1,2} =  \MatLog{\left(\Rotation_{2,w} \Rotation_{1,w}^{T} \right)\Estimate{\Rotation}_{2,1}^{T}}$.
The estimates $\Estimate{\Transform}_{2,1}$, $\Matrix{\Sigma}_\text{vo}$ and $\Estimate{\Rotation}_{2,1}$, $\Matrix{\Sigma}_{\text{hn}}$ are provided by our classical estimator and the HydraNet network respectively.  

\Cref{tab:kitti_fusion_stats} summarizes the results when we perform this fusion - and \Cref{fig:kitti_topdown} shows the final effect on the trajectory for sequence \texttt{00}. Similar to \cite{2018_Peretroukhin_Deep} and \cite{Peretroukhin2018}, we found that fusing deep rotation regression with classical methods results in motion estimates that significantly out-perform other methods that rely on deep regression alone. However, we note that even with consistent estimates, a small bias can affect the final fused estimates (e.g., sequence \texttt{05}) and removing bias is an important avenue for future work. Further, the KITTI dataset contains few deleterious effects that negatively affect classical algorithms, and therefore we expect that this fusion would produce even more pronounced improvements on more varied visual data.

\section{Conclusion}
In summary, we presented a method to regress probabilistic estimates of rotation using a deep multi-headed network structure. We used the quaternionic metric on $\LieGroupSO{3}$ to define a rotation average, and extracted anisotropic covariances by modelling uncertainty through noise injection on the manifold.
Further avenues for future work include obviating the need for supervised training by embedding the HydraNet structure within a Bayesian filter (see for example, \cite{Haarnoja2016-ph}), applying a HydraNet $\LieGroupSO{3}$ regression to improve convergence in non-convex pose graphs, and using HydraNet outputs to improve direct keyframe-based visual localization within a tight optimization loop.

\newpage
\appendix

\section{Rotation averaging}
The three different rotation metrics can be related to the angular (or geodesic) metric, $\RotDist{ang}$, as follows,

\begin{align}
\RotDist{ang}(\Matrix{R}_a, \Matrix{R}_b) &= \Norm{\MatLog{\Matrix{R}_a \Matrix{R}_b^T}}_2 
\\
 &= \theta,	
\\
\RotDist{quat}(\quat_a, \quat_b) &= \min\left( \Norm{\quat_a - \quat_b}_2, \Norm{\quat_a + \quat_b}_2 \right) 
\\ &= 2 \sin{\frac{\theta}{4}},	
\\
\RotDist{ang}(\Matrix{R}_a, \Matrix{R}_b) &= \Norm{\Matrix{R}_a - \Matrix{R}_b}_\mathrm{Frob} \\	 
&= 2\sqrt{2} \sin{\frac{\theta}{2}}.
\end{align}

Given a set of rotations parametrized by unit quaternions $\{\quat_i\}_{i=1}^n$,
\begin{equation}
\Mean{\quat} = \frac{\sum_{i=1}^n \quat_i}{\Norm{\sum_{i=1}^n \quat_i}},	
\end{equation}
solves
\begin{equation}
\quat = \ArgMin{\Matrix{R}(\quat) \in \LieGroupSO{3}} \sum_{i=1}^n \RotDist{quat}(\quat_i, \quat)^2	,
\end{equation}
so long as $\RotDist{ang}(\Matrix{R}(\Mean{\quat}), \Matrix{R}(\quat_i)) < \pi/2$. See \cite{Hartley2013-rc} for more details.

\section{Experiments}
\subsection{One-dimensional regression}

\begin{figure*}
	\centering
	\includegraphics[width=0.98\textwidth]{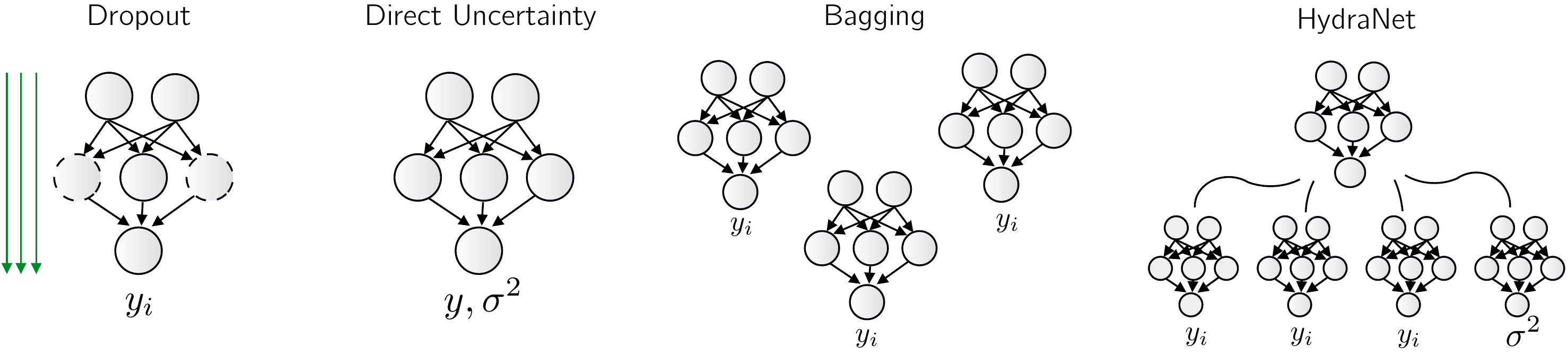}
	\vspace{-.5em}
	\caption{Different scalable approaches to neural network uncertainty. }
	\label{fig:nn-uncertainty}
\end{figure*}

For each uncertainty extraction, we used a four layer neural network (with 20 units per layer) with a Scaled Exponential Linear Unit (SELU). For the dropout method, we added dropout layers (with a small dropout probability, $p=0.03$, to account for the small network size as recommended by \cite{Gal2016-ny}). We performed 50 forward passes through the network, and computed the mean and variance of the outputs to determine the prediction and uncertainty estimate. For the ensemble bootstrap method, we trained ten separate models on bootstrapped samples of the training data. For HydraNet, we used the first two layers as the body, and branched the final two layers into ten heads.  One additional head was created that directly regressed an uncertainty estimate.

Every model in this experiment was trained for 3000 epochs using stochastic gradient descent with momentum, using minibatch sizes of 50 (refer to \Cref{tab:1d-hyp} for specific hyper-parameters). We repeated training 100 times, and recorded the test-time negative log likelihood for each method at each repetition.

\begin{figure}
	\centering
	\includegraphics[width=0.48\textwidth]{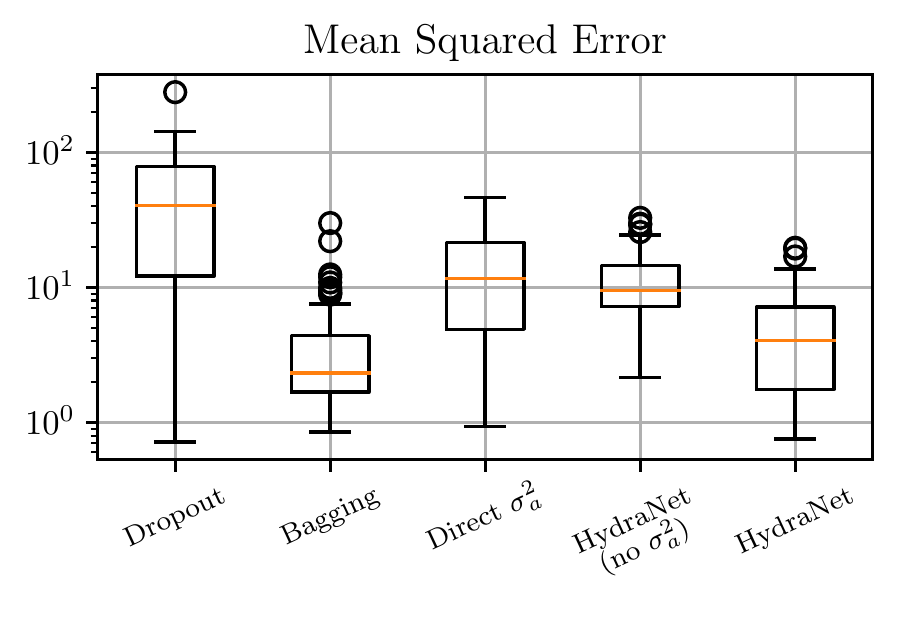}
	\caption{Mean squared errors for the different probabilistic regression models in 1D.}
	\label{fig:1d-mse}
\end{figure}

\begin{figure}
	\centering
	\includegraphics[width=0.48\textwidth]{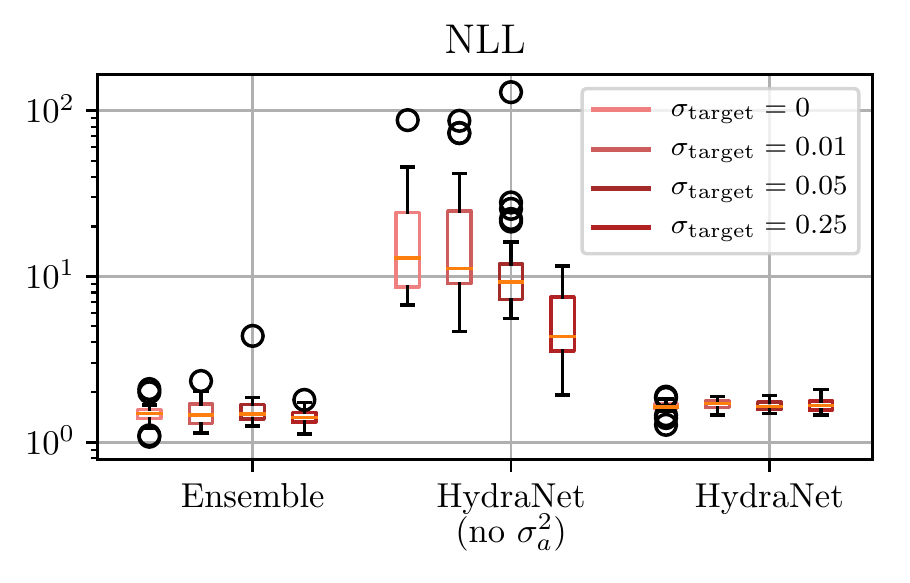}
	\caption{For the 1D experiment, we experimented with adding zero-mean Gaussian additive noise to the regression targets in an attempt to promote diversity amongst the outputs. We found that while this improved the uncertainty estimates gleaned from the HydraNet heads alone (what we call epistemic uncertainty) it made little difference once we included aleatoric uncertainty.}
	\label{fig:1d-target-noise}
\end{figure}

We present three additional figures here that were not included in the main paper. \Cref{fig:1d_uncertainty} presents four representative samples from the 100 repetitions for each method, and \Cref{fig:1d-mse} presents mean squared errors for each method. The last figure, \Cref{fig:1d-target-noise}, details the effects of adding zero mean Gaussian noise to the regression targets during training. We experimented with this approach to try and promote more diversity amongst the HydraNet heads within training data. We found, however, that although this does improve the negative log likelihoods for HydraNet with only epistemic uncertainty (i.e., the sample variance over the head outputs), its benefits were non-existent for the full HydraNet approach. Namely, since the full HydraNet approach uses an NLL loss, the network tended to account for target noise by enlarging the aleatoric uncertainty rather than overfitting each head to a specific target.

\begin{figure*}[h!]
	\centering
	\begin{subfigure}[]{\textwidth}
		\includegraphics[width=\textwidth]{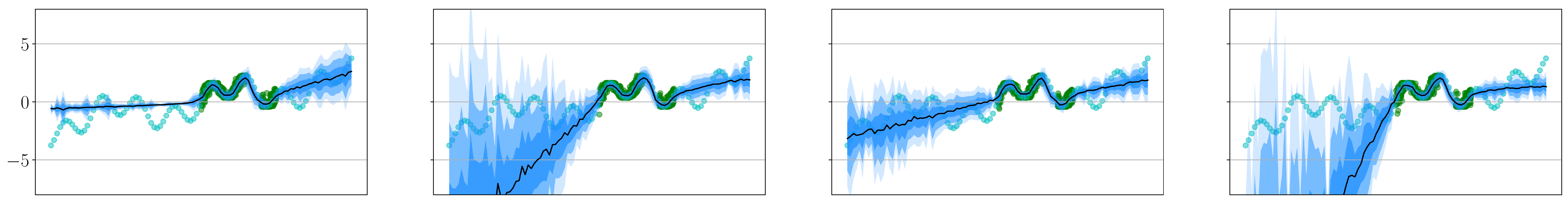}
		\caption{Dropout}
		\label{fig:1D_dropout}
	\end{subfigure}
	\hfill
	\begin{subfigure}[]{\textwidth}
		\includegraphics[width=\textwidth]{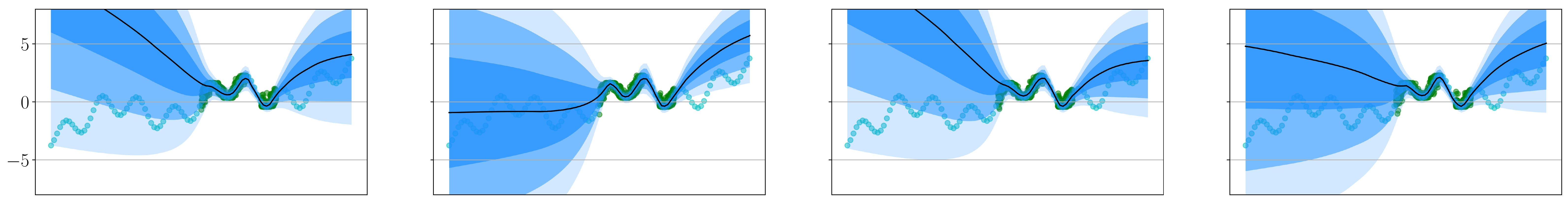}
		\caption{Direct Uncertainty}
		\label{fig:1D_direct}
	\end{subfigure}
	\hfill
	\begin{subfigure}[]{\textwidth}
		\includegraphics[width=\textwidth]{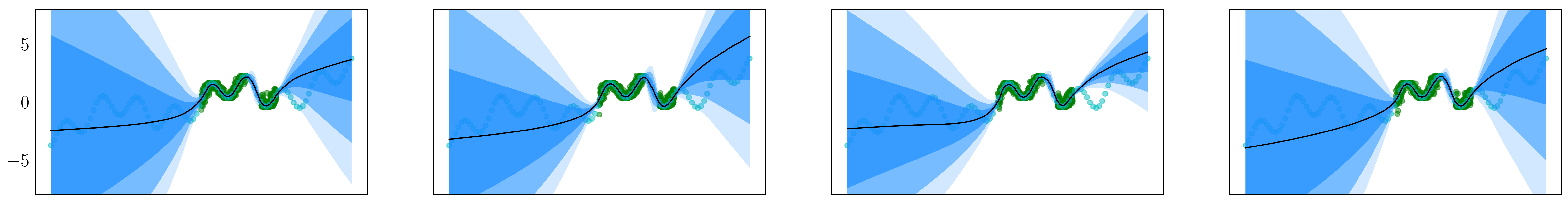}
		\caption{Bagging}
		\label{fig:1D_bagging}
	\end{subfigure}
	\hfill	
	\begin{subfigure}[]{\textwidth}
		\includegraphics[width=\textwidth]{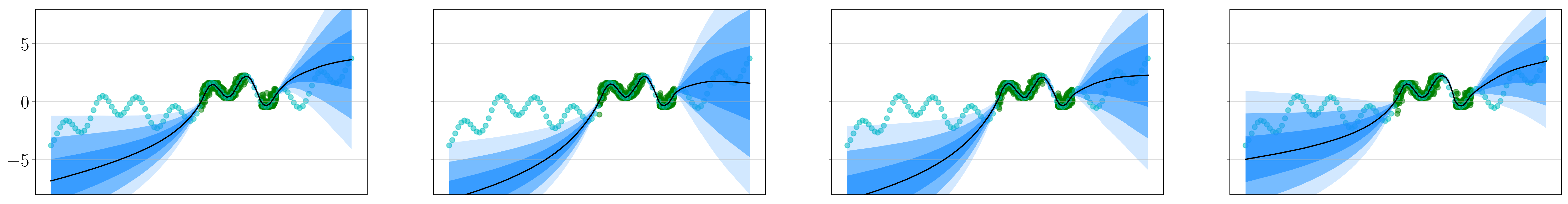}
		\caption{HydraNet (no direct uncertainty)}
		\label{fig:1D_hydranet_undirect}
	\end{subfigure}
	\hfill
	\begin{subfigure}[]{\textwidth}
		\includegraphics[width=\textwidth]{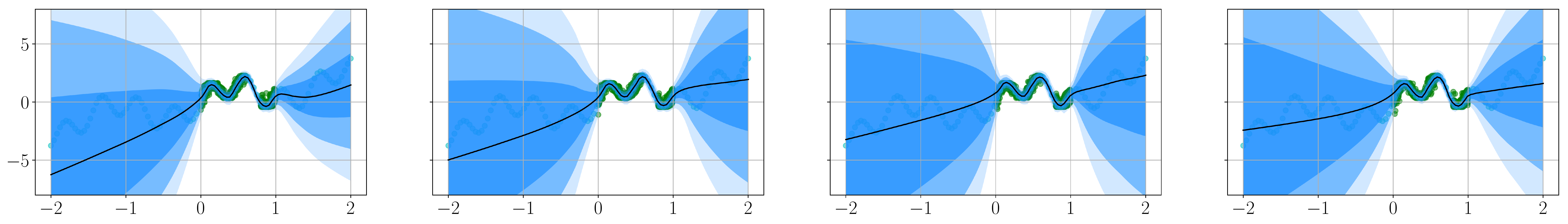}
		\caption{HydraNet}
		\label{fig:1D_hydranet}
	\end{subfigure}
	\hfill		
	\caption{A comparison of different ways to extract uncertainty from deep networks. Each shade of blue represents one standard deviation $\sigma$ produced by the model.}
	\label{fig:1d_uncertainty}
\end{figure*}

\begin{table*}[]
	\centering
	\caption{Hyper-parameters for 1D training.}
	\begin{threeparttable}
	\begin{tabular}{lccc}
		\toprule
		\textbf{Uncertainty Method} & \textbf{Learning Rate} & \textbf{Momentum} & \textbf{Dropout (\%)} \\ \midrule
		Dropout & 0.05 & 0.5 & 3 \\
		Direct Regression & 0.0001 & 0 & 0 \\
		Bagging & 0.01 & 0.9 & 0 \\
		HydraNet (no direct uncertainty) & 0.01 & 0.9 & 0 \\
		HydraNet & 0.01 & 0.1 & 0 \\ \bottomrule
	\end{tabular}
	\label{tab:1d-hyp}
\end{threeparttable}
\end{table*}

\subsection{Hemisphere world}

For this experiment, we created a synthetic world with a 6 $\times$ 6 grid of landmarks, each spaced one meter apart. Our monocular camera resided on a hemisphere (of radius 25 meters) from the centre of the landmark grid. The camera sensor was 500 $\times$ 500 pixels, with a principal point in the middle of the sensor and a focal length of 500 pixels.  We added zero-mean Gaussian noise of unit pixel variance to each landmark projection. 

The network consisted of five residual blocks, each containing a fully connected layer and a ReLU non-linearity. For each camera location, we projected all 36 landmarks onto the image plane, added noise, and then stored 72 image coordinates as training or test input. 

\subsection{7-Scenes}

\begin{figure*}[h!]
	\centering
	\begin{subfigure}[]{0.33\textwidth}
		\includegraphics[width=\textwidth]{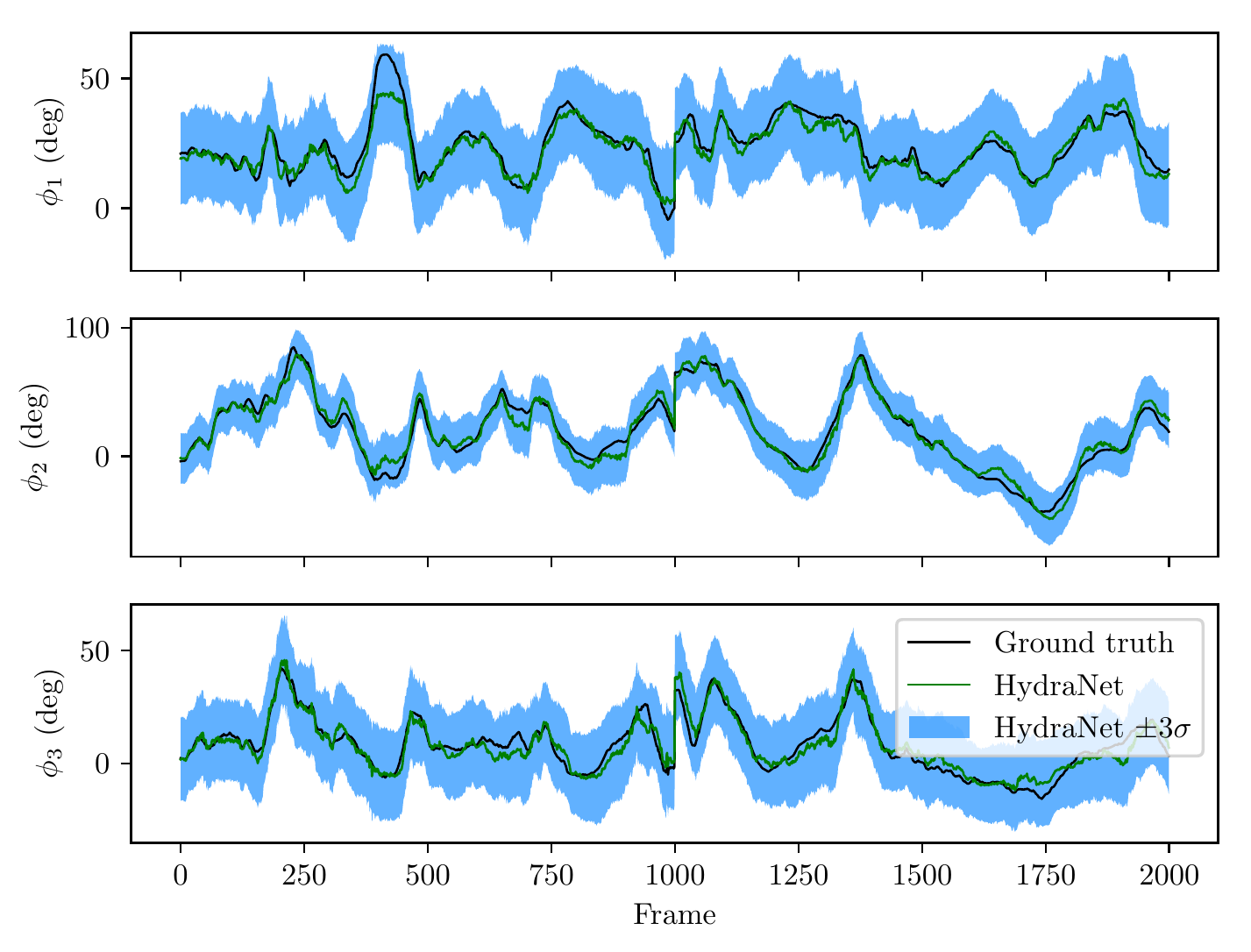}
		\caption{Chess}
	\end{subfigure}
	\begin{subfigure}[]{0.33\textwidth} 
		\includegraphics[width=\textwidth]{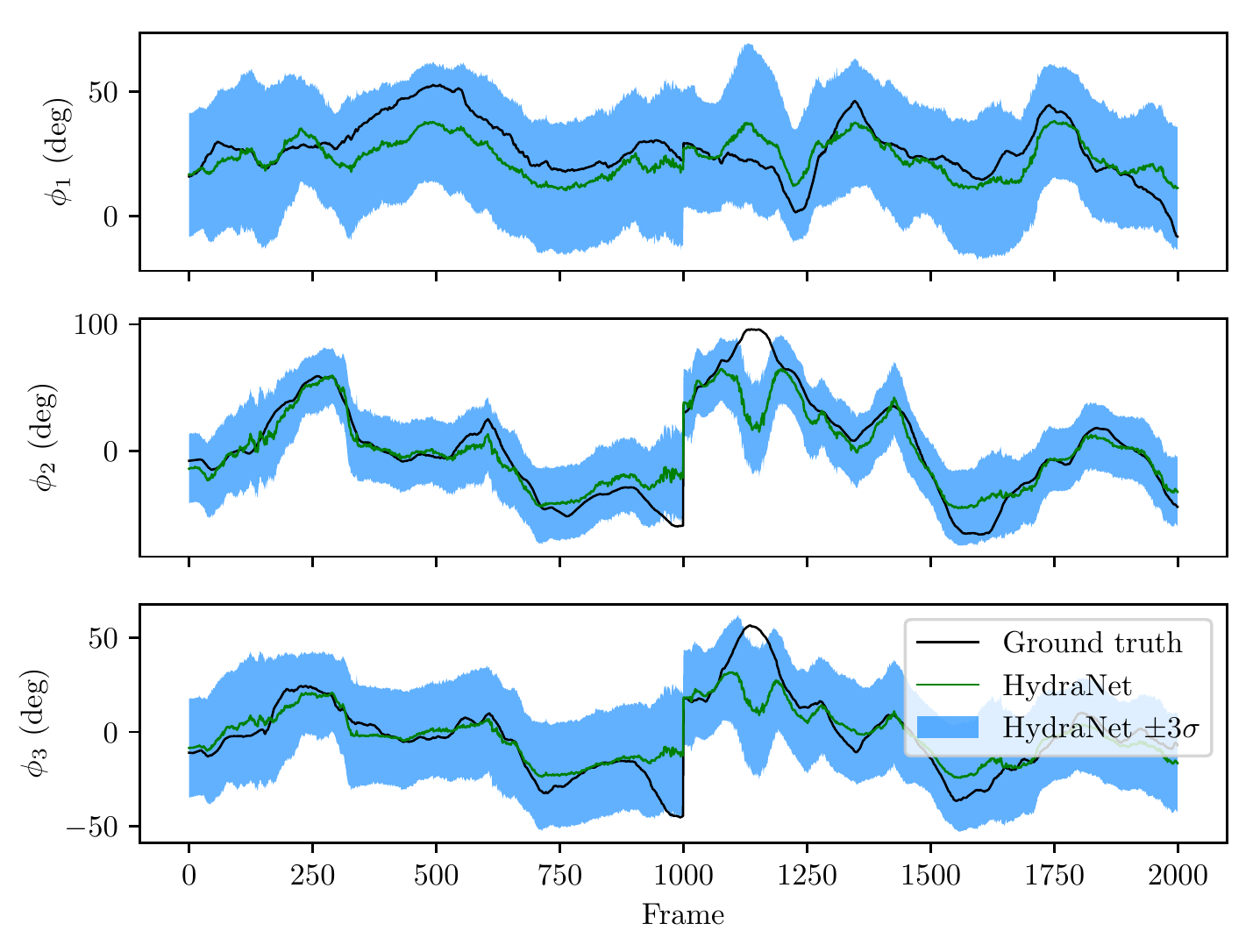}
		\caption{Fire}
	\end{subfigure}
	\begin{subfigure}[]{0.33\textwidth} 
		\includegraphics[width=\textwidth]{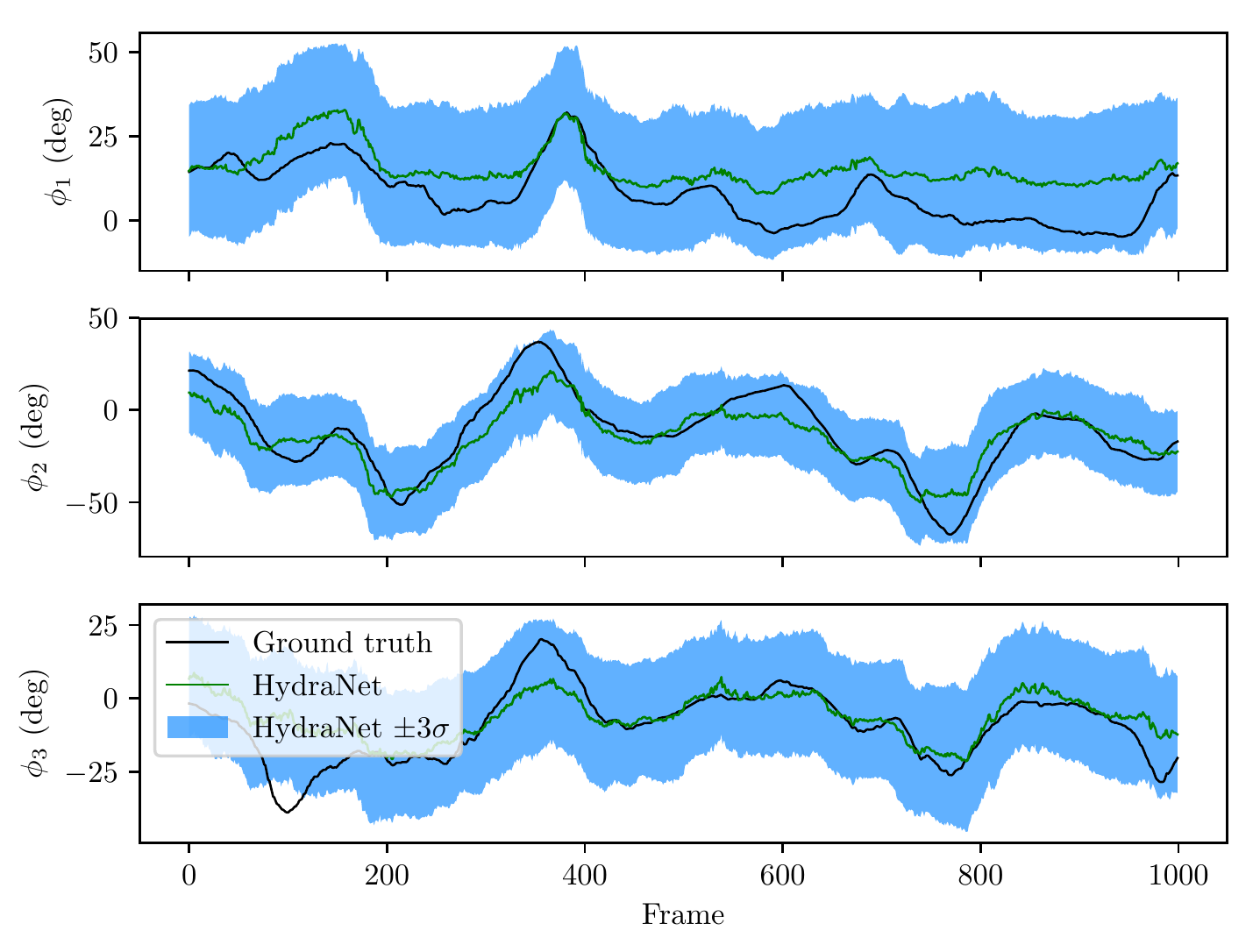}
		\caption{Heads}
	\end{subfigure}
	\begin{subfigure}[]{0.33\textwidth} 
		\includegraphics[width=\textwidth]{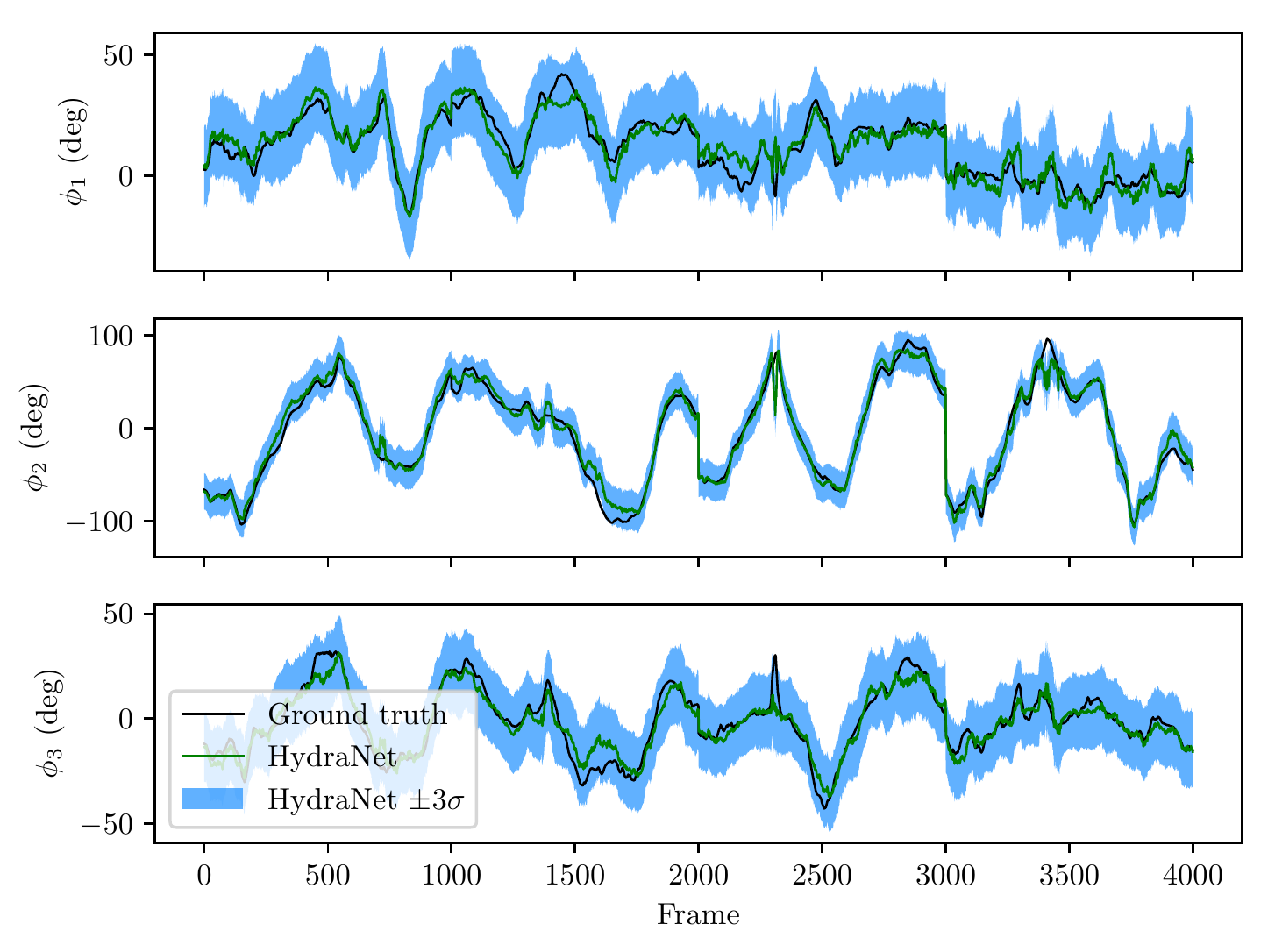}
		\caption{Office}
	\end{subfigure}
	\begin{subfigure}[]{0.33\textwidth} 
		\includegraphics[width=\textwidth]{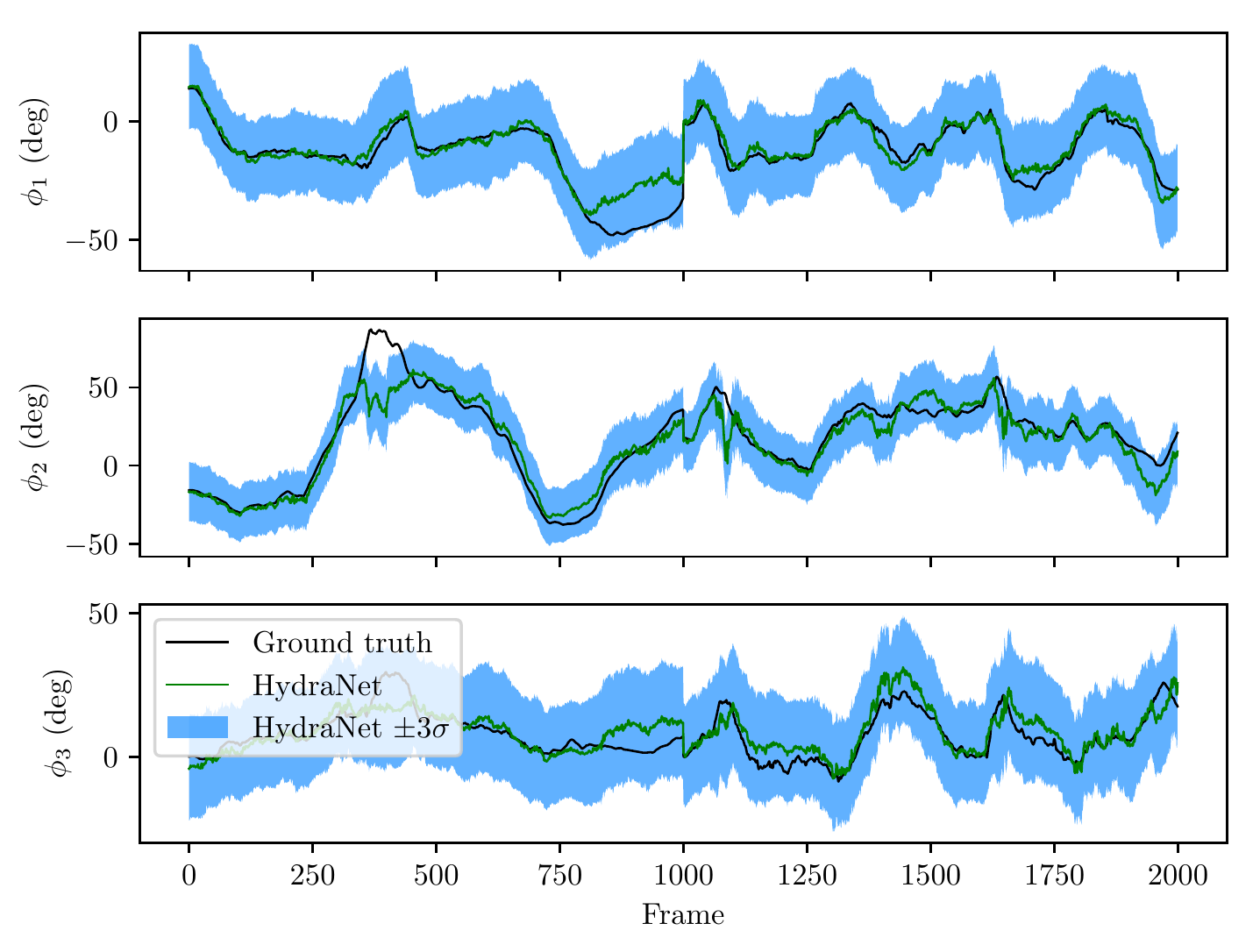}
		\caption{Pumpkin}
	\end{subfigure}
	\begin{subfigure}[]{0.33\textwidth} 
		\includegraphics[width=\textwidth]{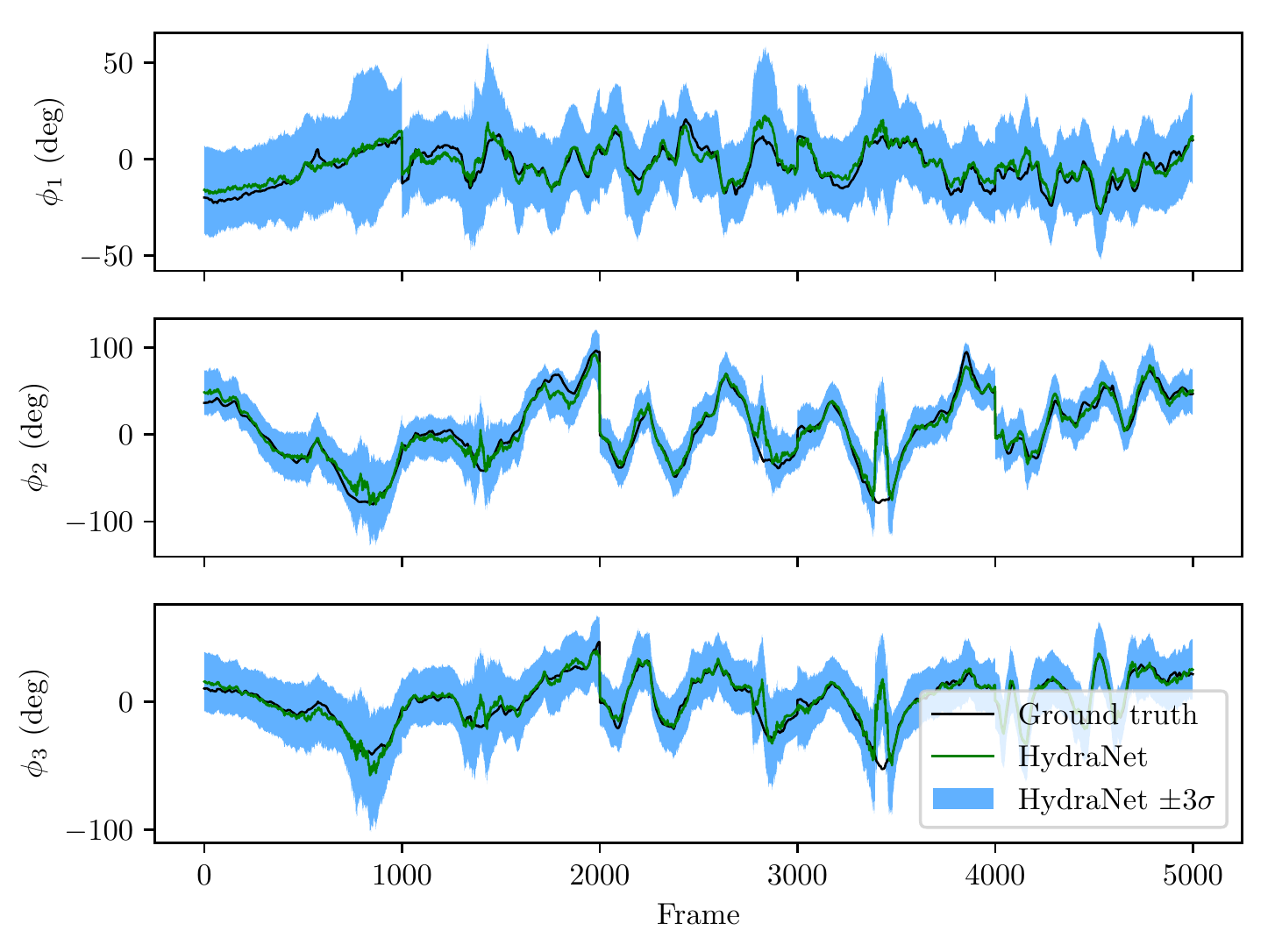}
		\caption{Kitchen}
	\end{subfigure}
	\begin{subfigure}[]{0.33\textwidth} 
		\includegraphics[width=\textwidth]{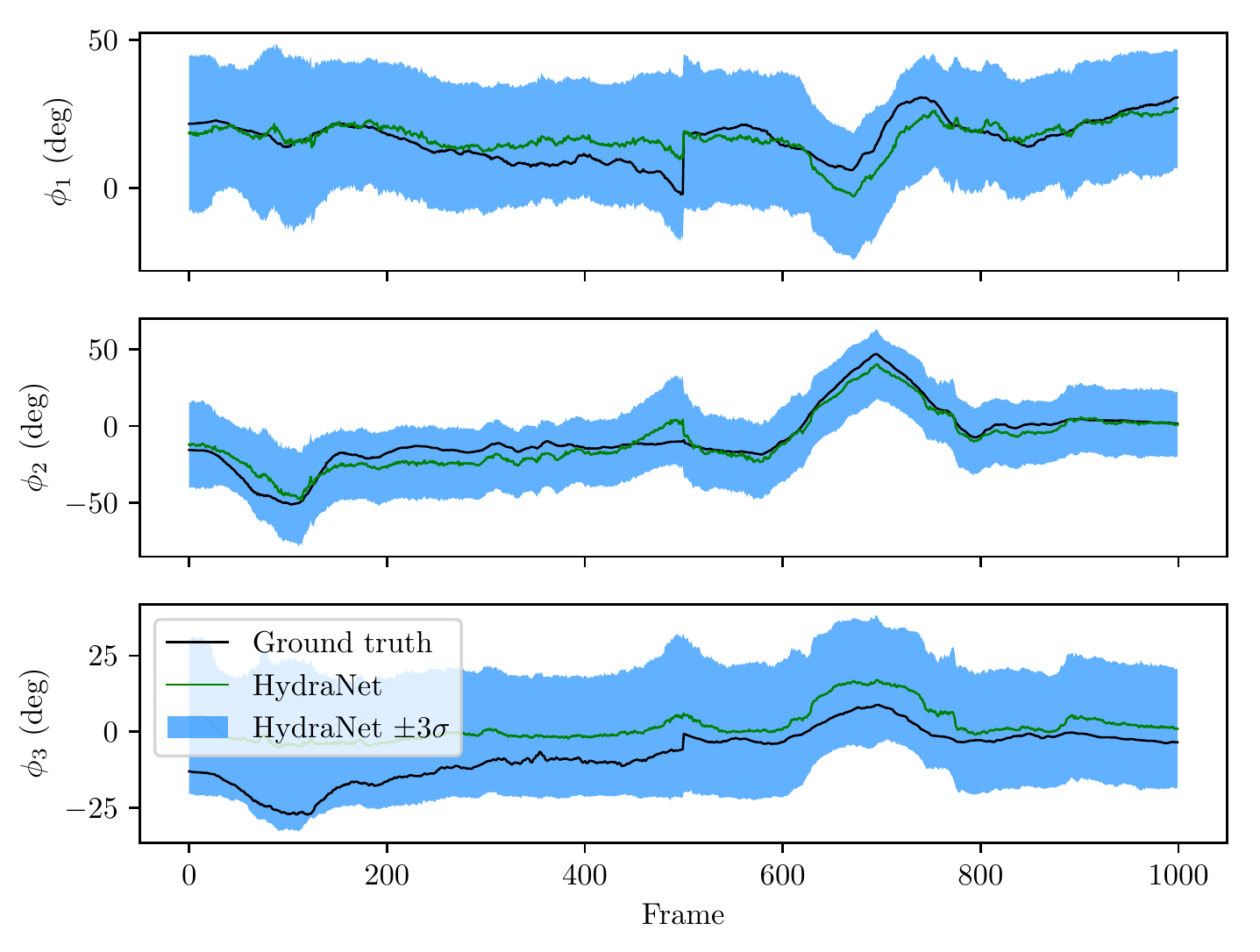}
		\caption{Stairs}
	\end{subfigure}
	
	\caption{Probabilistic regression plots for all seven datasets from the 7-Scenes dataset.}
	\label{fig:7scenes_regression}
\end{figure*}

\Cref{fig:7scenes_regression} presents regression results on all seven scenes from the 7-scenes dataset. Our model consisted of a \texttt{resnet34} body (pre-trained, but not frozen) with 25+1 heads in the same structure as the synthetic experiment. We used the Adam optimizer with a learning rate of $5 \times 10^{-5}$ for all scenes, and trained each model for 15 epochs, selecting the one with the lowest negative log likelihood.

\subsection{KITTI}
\subsubsection{Network details}
Our custom convolutional network was built using PyTorch as follows:

\begin{python}
self.cnn = torch.nn.Sequential(
    conv_unit(2, 64),
    conv_unit(64, 128),
    conv_unit(128, 256),
    conv_unit(256, 512),
    conv_unit(512, 1024),
    conv_unit(1024, 1024),
    conv_unit(1024, 1024)
)
\end{python}
with each \texttt{conv\_unit} defined as,
\begin{python}
def conv_unit(in, out, ks=3, st=2, pad=1):
        return torch.nn.Sequential(
            torch.nn.Conv2d(in, out, 
            	kernel_size=ks, 
            	stride=st, 
            	padding=pad),
            torch.nn.BatchNorm2d(out),
            torch.nn.ReLU()
        )	
\end{python}
and the head structure being identical to both of the previous experiments. Our two-dimensional flow image was constructed using OpenCV with the function \texttt{calcOpticalFlowFarneback()} from two RGB images converted to grayscale. We trained the network using the Adam optimizer, with a learning rate of $5 \times 10^{-5}$ and no pre-training. We found that augmenting the dataset with rotation targets and inputs that represented both the forward and reverse temporal pairs improved generalization. 
%
%
%

\begin{figure*}[h!]
	\centering
	\begin{subfigure}[]{0.33\textwidth}
		\includegraphics[width=\textwidth]{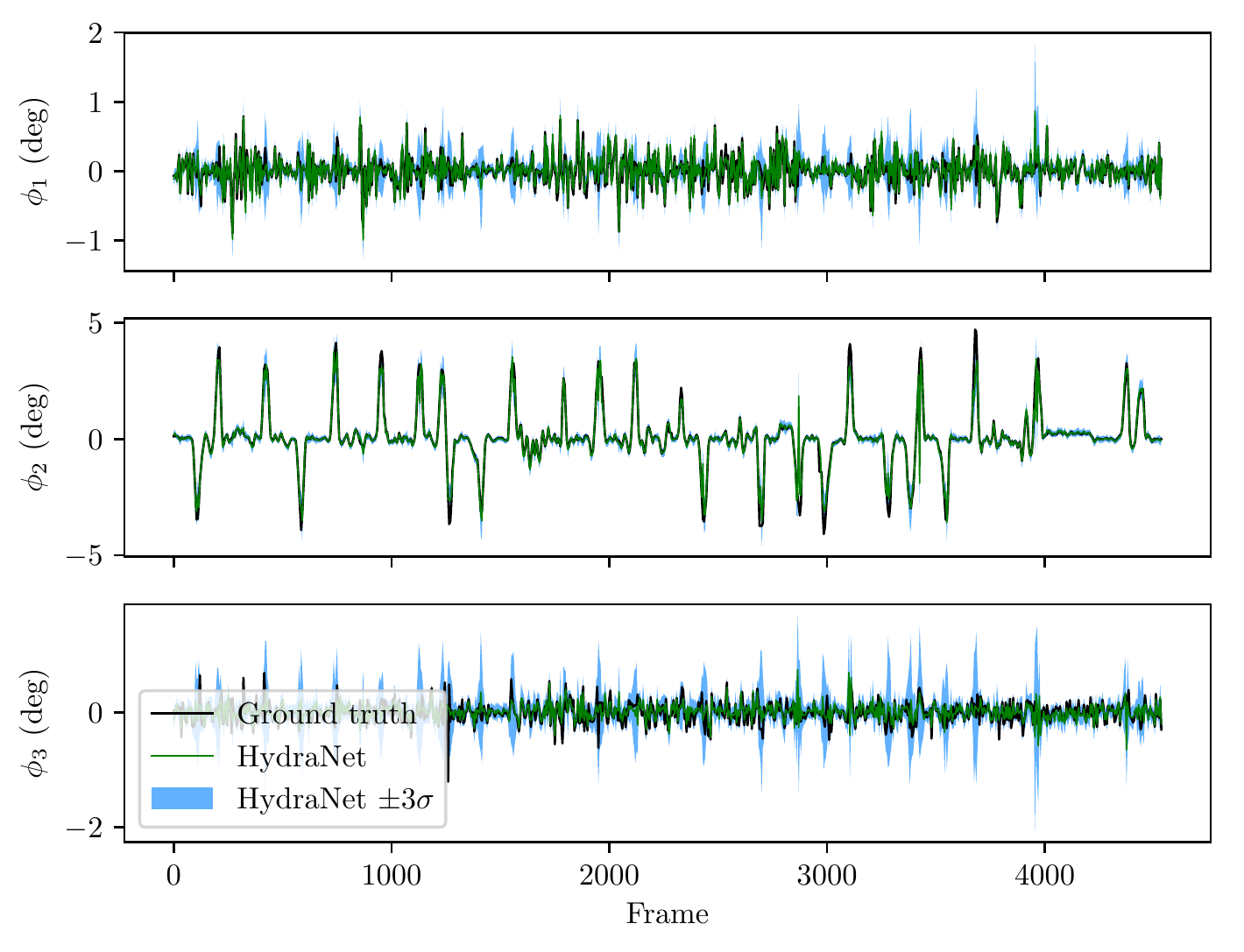}
		\caption{\texttt{00}}
	\end{subfigure}
	\begin{subfigure}[]{0.33\textwidth} 
		\includegraphics[width=\textwidth]{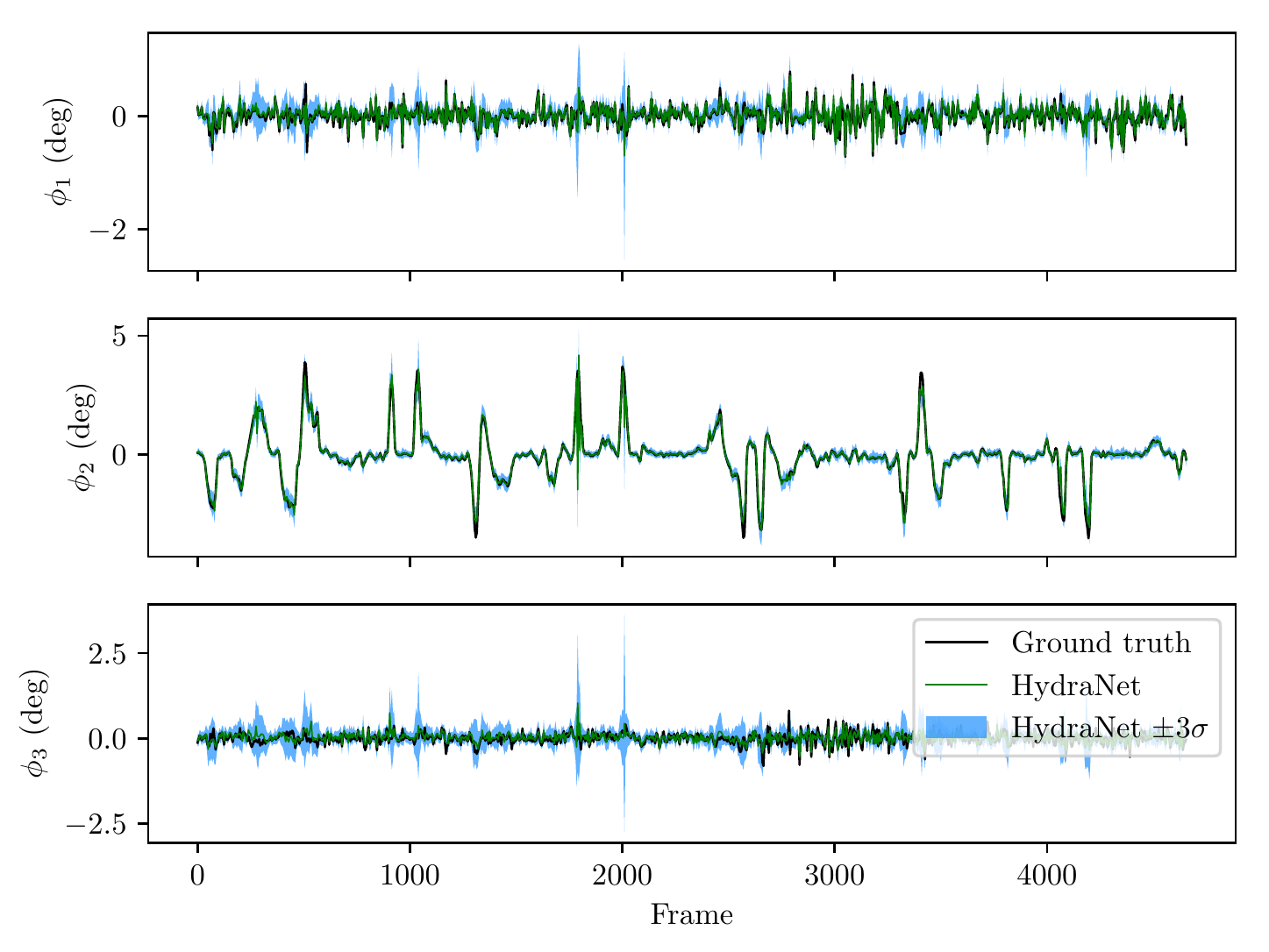}
		\caption{\texttt{02}}
	\end{subfigure}
	\begin{subfigure}[]{0.33\textwidth} 
		\includegraphics[width=\textwidth]{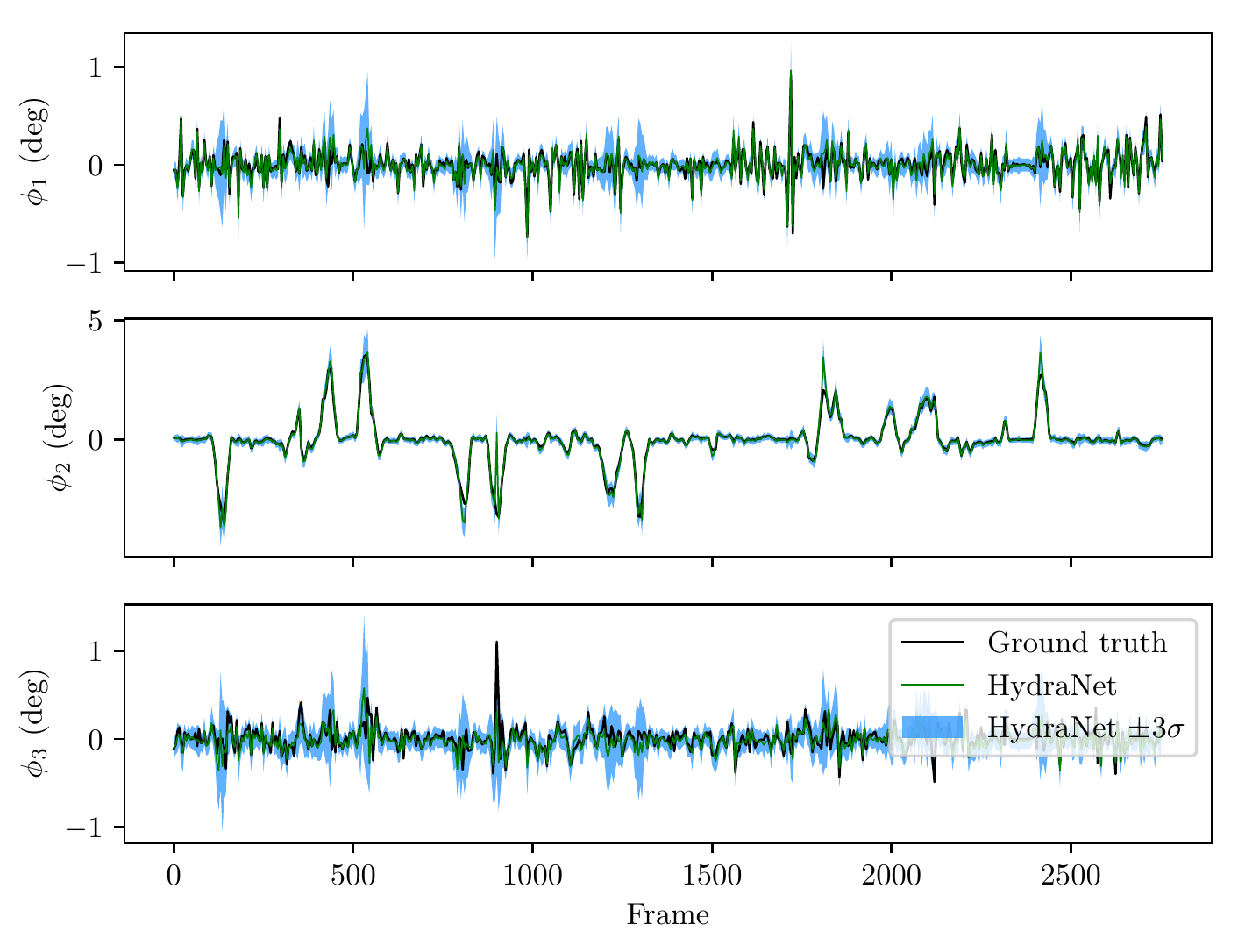}
		\caption{\texttt{05}}
	\end{subfigure}
	\begin{subfigure}[]{0.33\textwidth} 
		\includegraphics[width=\textwidth]{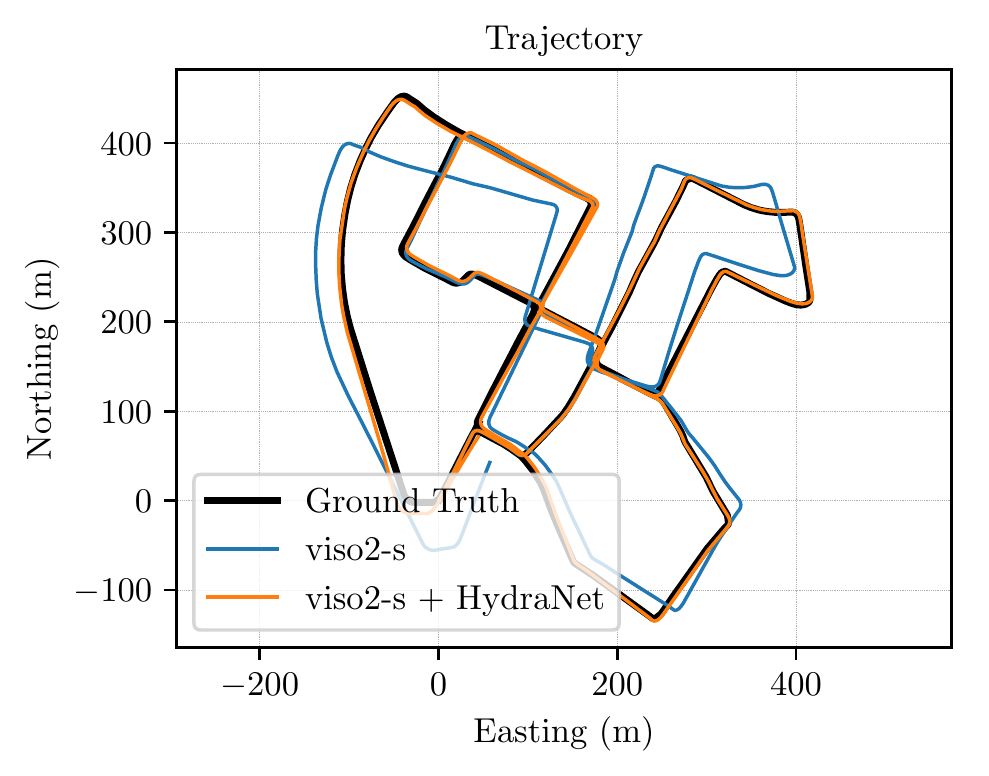}
		\caption{\texttt{00}}
	\end{subfigure}
	\begin{subfigure}[]{0.33\textwidth} 
		\includegraphics[width=\textwidth]{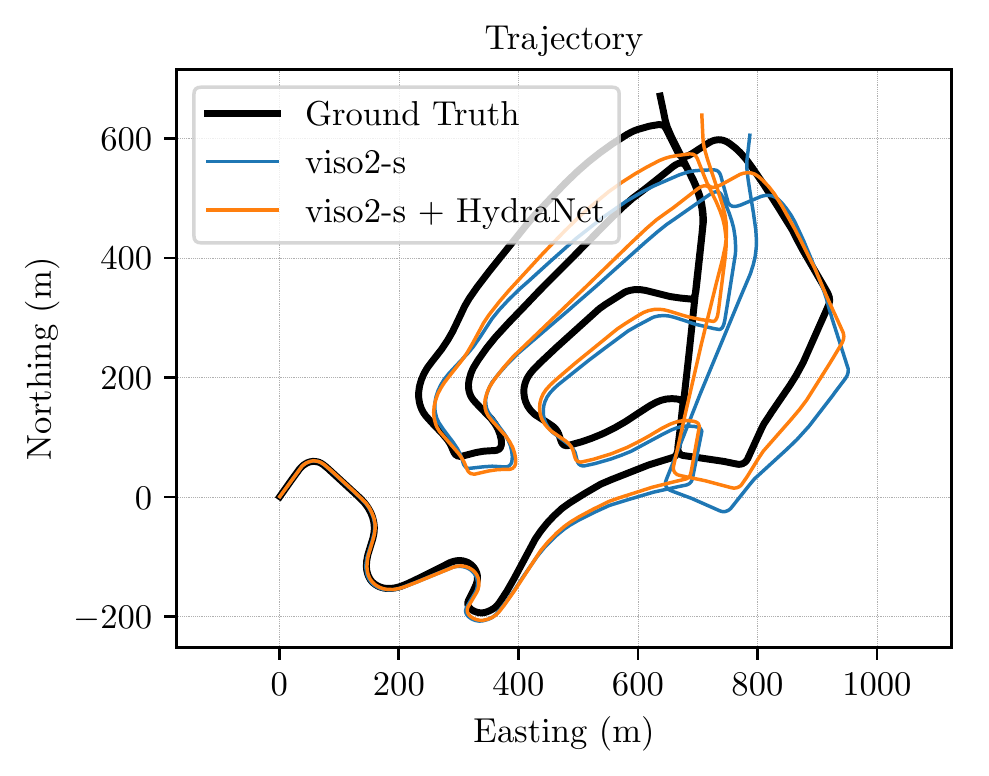}
		\caption{\texttt{02}}
	\end{subfigure}
	\begin{subfigure}[]{0.33\textwidth} 
		\includegraphics[width=\textwidth]{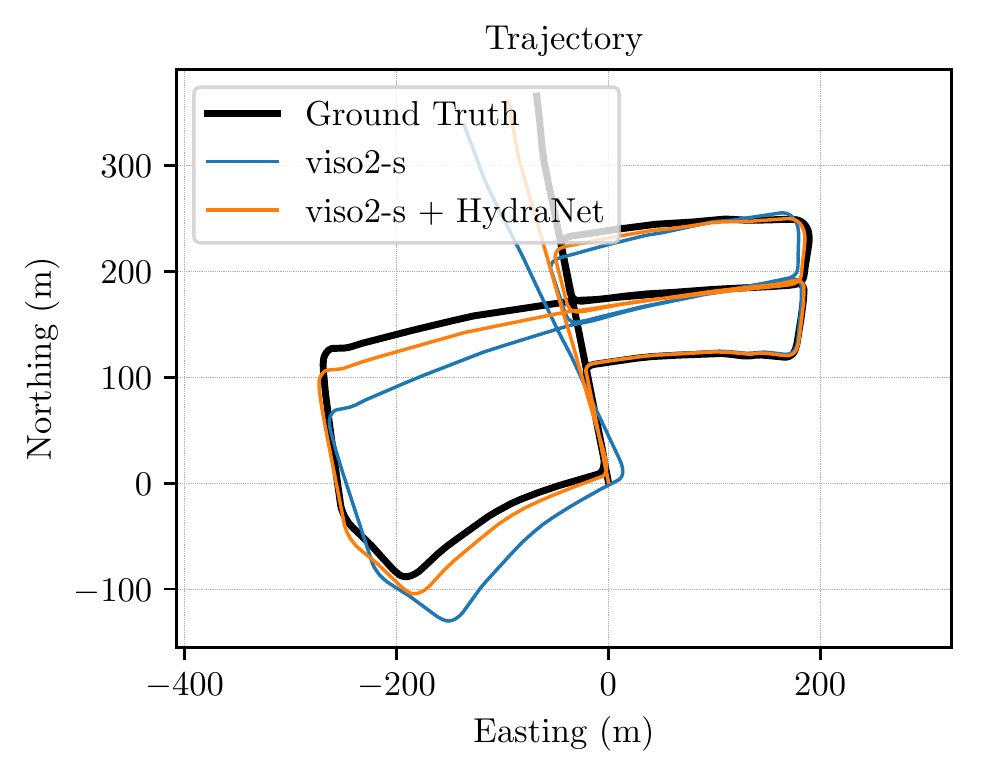}
		\caption{\texttt{05}}
	\end{subfigure}
	\caption{KITTI frame-to-frame rotation probabilistic regression for sequences \texttt{00}, \texttt{02} and \texttt{05}. Top-down trajectory plots show localization improvements after fusion with a classical stereo visual odometry pipeline.}
	\label{fig:kitti_regression}
\end{figure*}

{\small
\bibliographystyle{ieee}
\bibliography{iccv_refs}
}

\end{document}